\documentclass[pdflatex,sn-apa]{sn-jnl}% Basic Springer Nature Reference Style/Chemistry Reference Style
\usepackage{graphicx}%
\usepackage{multirow}%
\usepackage{amsmath,amssymb,amsfonts, bm}%
\usepackage{bbm, dsfont}
\usepackage{amsthm}%
\usepackage{mathrsfs}%
\usepackage[title]{appendix}%
\usepackage{xcolor}%
\usepackage{textcomp}%
\usepackage{manyfoot}%
\usepackage{booktabs}%
\usepackage{algorithm}%
\usepackage{algorithmicx}%
\usepackage{algpseudocode}%
\usepackage{listings}%
\usepackage{paralist}
\usepackage{tabularx,longtable}
\usepackage{hyperref}
\newtheoremstyle{thmstyleone}% Numbered
{18pt plus2pt minus1pt}% Space above
{18pt plus2pt minus1pt}% Space below
{\small\itshape}% Body font
{0pt}% Indent amount
{\small\bfseries}% Theorem head font
{}% Punctuation after theorem head
{.5em}% Space after theorem headi
{\thmname{#1}\thmnumber{\@ifnotempty{#1}{ }\@upn{#2}}%
  \thmnote{ {\the\thm@notefont(#3)}}}% Theorem head spec (can be left empty, meaning `normal')
\newtheorem{theorem}{Theorem}%  meant for continuous numbers
\DeclareMathOperator*{\argmin}{arg\,min}
\begin{document}
%%%%%%%%%%%%%%%%%%%%%%%%%%%%%%%%%%%%%%%%%%%%%%%%%%%%%%%%%%%%%%%%%%
%%%%%%%%%%%%%%%%%%%%%%%%%%%%%%%%%%%%%%%%%%%%%%%%%%%%%%%%%%%%%%%%%%
\title[Article title]{Conformal Approach To Gaussian Process Surrogate Evaluation With Coverage Guarantees}

\author[1,3,4]{\fnm{Edgar}\sur{Jaber}}\email{edgar.jaber@edf.fr}\equalcont{These authors contributed equally to this work.}
\author*[2,3,8]{\fnm{Vincent}\sur{Blot}}\email{vblot@quantmetry.com}\equalcont{These authors contributed equally to this work.}
\author[2,5]{\fnm{Nicolas}\sur{Brunel}}
\author[1,6]{\fnm{Vincent}\sur{Chabridon}}
\author[1]{\fnm{Emmanuel}\sur{Remy}}
\author[1,6,7]{\fnm{Bertrand}\sur{Iooss}}
\author[3]{\fnm{Didier}\sur{Lucor}}
\author[4,5]{\fnm{Mathilde}\sur{Mougeot}}
\author[3,8]{\fnm{Alessandro}\sur{Leite}}

\affil[1]{\orgname{EDF R\&D}, \orgaddress{\street{6 Quai Watier}, \postcode{78401}, \city{Chatou}, \country{France}}}
\affil*[2]{\orgname{Quantmetry}, \orgaddress{\street{52 Rue d'Anjou}, \postcode{75008}, \city{Paris}, \country{France}}}
\affil[3]{\orgname{Paris-Saclay University, CNRS, Laboratoire Interdisciplinaire des Sciences du Numérique}, \orgaddress{\postcode{91405}, \city{Orsay}, \country{France}}}
\affil[4]{\orgname{Paris-Saclay University, CNRS, ENS Paris-Saclay, Centre Borelli}, \orgaddress{\postcode{91190}, \city{Gif-sur-Yvette}, \country{France}}}
\affil[5]{\orgname{ENSIIE}, \orgaddress{\street{1 square de la Résistance}, \postcode{91000}, \city{Évry-Courcouronnes}, \country{France}}}
\affil[6]{\orgname{SINCLAIR AI Lab.},  \orgaddress{\city{Saclay}, \country{France}}}
\affil[7]{\orgname{Institut de Mathématiques de Toulouse},  \orgaddress{\street{118 route de Narbonne}, \postcode{31062}, \city{Toulouse}, \country{France}}}
\affil[8]{\orgname{TAU, INRIA}, \orgaddress{\street{Bat. 660 Claude Shannon, Paris-Saclay University}, \postcode{91405}, \city{Orsay}, \country{France}}}

%%==================================%%
%% Abstract %%
%%==================================%%
\abstract{
 Gaussian processes (GPs) are a Bayesian machine learning approach widely used to construct surrogate models for the uncertainty quantification of computer simulation codes in industrial applications. It provides both a mean predictor and an estimate of the posterior prediction variance, the latter being used to produce Bayesian credibility intervals. Interpreting these intervals relies on the Gaussianity of the simulation model as well as the well-specification of the priors which are not always appropriate. We propose to address this issue with the help of conformal prediction. In the present work, a method for building adaptive cross-conformal prediction intervals is proposed by weighting the non-conformity score with the posterior standard deviation of the GP. The resulting conformal prediction intervals exhibit a level of adaptivity akin to Bayesian credibility sets and display a significant correlation with the surrogate model local approximation error, while being free from the underlying model assumptions and having frequentist coverage guarantees. These estimators can thus be used for evaluating the quality of a GP surrogate model and can assist a decision-maker in the choice of the best prior for the specific application of the GP. The performance of the method is illustrated through a panel of numerical examples based on various reference databases. Moreover, the potential applicability of the method is demonstrated in the context of surrogate modeling of an expensive-to-evaluate simulator of the clogging phenomenon in steam generators of nuclear reactors.
}


%%==================================%%
%% Title %%
%%==================================%%
\maketitle

%%==================================%%
\section*{Acknowledgments}
%%==================================%%
The PhD programs of the contributing authors are funded by the French National Association for Technological Research (ANRT) under grants n. 2022/1412, 2022/0667 respectively. Part of this work was supported by EDF and Quantmetry, and initiated during the Summer Mathematical Research Center on Scientific Computing (whose French acronym is ``CEMRACS''), which took place at CIRM, in Marseille (France), from July 17 to August 25, 2023 (\url{http://smai.emath.fr/cemracs/cemracs23/index.html}). The first contributing author would like to thank Morgane Garo Sail and Charlotte G\'ery (project managers at EDF R\&D) for the financial and organizational supports through their projects.
%%==================================%%
\section{Introduction}
\label{sec1}
%%==================================%%

%%-----------------------------%%
\subsection{Motivation and overview}
%%-----------------------------%%

In the field of the analysis and design of computer experiments \citep{fang_li_sudjianto_book_2005}, the so-called ``Verification, Validation and Uncertainty Quantification'' (known as ``VV\&UQ'') framework now became a gold-standard in many engineering fields in order to rigorously assess the impact of various sources of uncertainty affecting some input variables of numerical simulation models \citep{deRocquigny_devictor_tarantola_book_2008, sullivan_book_2015, springer_handbook_of_uq_2017}. To be more specific, uncertainty quantification (UQ) relies on the definition of a computer model as a function $g$ which maps a $d$-dimensional input vector $X \in \mathcal{X} \subseteq \mathbb{R}^{d}$ to an output variable of interest, typically scalar $Y \in \mathcal{Y}\subseteq \mathbb{R}$, by the input-output relationship $Y=g(X)$. These computer models play a key role in engineering as they are used for decision-making regarding the management of industrial assets, with various applications from maintenance scheduling to risk assessment of critical systems. Typically, $g$ can be a numerical solver for a system of partial differential equations, or a high-fidelity multiphysics model. In this methodology, uncertainties are often treated in a probabilistic way, which means that one can draw samples according to the joint probability distribution of $X$ in order to propagate the input uncertainties (typically, using Monte Carlo sampling techniques, see, e.g., \citep{Rubinstein_Kroese_book_1981}) through the computer model $g$ in order to get the output empirical distribution. During this step, the code $g$ is often considered as a ``black-box'', meaning that this procedure is completely non-intrusive with respect to the code. As soon as $g$ is expensive-to-evaluate (e.g., from one hour to several days for a single call), standard UQ procedures often become computationally intractable.

In this context, a so-called ``metamodel'' (also called ``surrogate model''), denoted by $\widehat{g}$, can be constructed to reduce the computational burden. The building strategy of such metamodels require to choose an input design of experiments (DoE) for sampling $X$ and a family of metamodels. These choices are often guided by prior input probabilistic modeling, computational cost constraints or potential prior knowledge about the regularity of the computer code. Once the input probability distribution is given, several techniques can be used to get the DoE, from basic Monte Carlo sampling to more advanced space-filling techniques (see, e.g., \citep{fang_li_sudjianto_book_2005}). Then, the input-output DoE is used to train and validate the metamodel $\widehat{g}$ in a similar fashion as what is currently done in supervised machine learning (ML). Similar to ML algorithms, a large panel of metamodels is available in the literature and many of them are used in industrial practice \citep{cheng_2023}. Some of them are related to deterministic approximation methods (e.g., polynomial chaos expansions \citep{sudret_2014, elmocayd_et_al_2018, jaber_et_al_2023} or artificial neural networks \citep{tripathy_et_al_2018, lefebvre_et_al_2023}), i.e., they may optimally account for the probabilistic distribution of the inputs but do not model their discrepancy with respect to the true response with the help of stochastic quantities. Others are probabilistic by nature (typically, Gaussian processes or GPs \citep{rasmussen_williams_book_2006}) and  naturally equipped with stochastic approximations of their errors but blind to the underlying distributions of the inputs (except through the DoE). 
Once a metamodel is built and used as a surrogate of the original code, it serves either to speed up critical decision-making processes or to facilitate intensive uncertainty analyses which are useful for global sensitivity analysis or code calibration (with respect to experimental data). The choice of a type of surrogate is driven by various considerations such as the size of the training DoE, the dimension of the input and output spaces or the regularity of the output with respect to the inputs. The present paper focuses primarily on GPs. 

A crucial step in surrogate modeling is required to evaluate the ``quality'' of a metamodel. Typically, in GP regression (also known as ``Kriging''), various validation metrics have been developed in the last decades in order to assess the predictive quality of the fitted GP metamodel \citep{demay_et_al_2022,wiecro23,marioo23a}. Some effort has been put to propose validation metrics enabling to go beyond the measure of the quality of the mean prediction (typically measured by the predictivity coefficient), for instance by measuring the quality of the posterior predicted variance. As proposed in \citep{deCarvalho_et_al_22} and \citep{jaber_et_al_2023}, additional cross-validation (such as $K$-fold, or LOO) techniques can be used for assessing the robustness of the estimation on these validation indicators. 

However, to the best of the authors' knowledge, such a topic is still an open question and no strong consensus has been reached regarding the metrics that should be used in general for validating a GP metamodel or any other metamodel in general. Another crucial ingredient of an efficient surrogate model should be its adaptivity to the level of certainty of the local information. By conditioning on observed data, GP model gains a better understanding of the underlying patterns and is less likely to be overly confident in regions with limited or no data. The reliability of GP predictions is a crucial aspect, especially when considering the Bayesian credibility intervals associated with those predictions. 

This reliability is influenced by the quality and quantity of the training data. In regions where there is more observed and less noisy data, the GP tends to be more confident and reliable in its predictions, leading to narrower credibility intervals. Properly tuned hyperparameters and chosen covariance kernels can also enhance the reliability of the GP, while poor choices may lead to overly optimistic or conservative uncertainty estimates. Along these lines, an alternative approach, proposed by \citep{acharki_robust_2023} aimed at enhancing the predictive capacity of a GP metamodel by optimizing the hyperparameters of the kernel in order to tackle model misspecification and to obtain more robust Bayesian credibility intervals.  However this method still heavily relies on the assumption of Gaussianity of the original model.

Calibration of the credibility intervals is also very important for assessing the reliability of uncertainty estimates. Indeed, well-calibrated intervals provide a true measure of the model's uncertainty, ensuring that, for a given confidence level, the actual surrogate predictions fall within the predicted intervals. 

In the present paper, another track is pursued: the idea relies on using conformal prediction (CP) methods \citep{vovk_gammerman_shafer_book_2005} to quantify the GP prediction uncertainty while not depending on any Gaussian assumption and well-specification of the posterior kernel, two key elements which are required to fully interpret the Bayesian credibility intervals. This complementary tool can thus be used to assist a decision-maker on evaluating the general quality of a GP surrogate, in the light of the application for which it is used.

CP has gained in the last decade a huge popularity within the ML community since it allows to perform distribution-free UQ in both classification and regression applications \citep{vovk_gammerman_shafer_book_2005,gentle_intro_angelopoulos_2023}. The CP paradigm enables the estimation of frequentist prediction intervals for any ML model (and, consequently, any metamodel) that are agnostic to the specific family of models used during the learning step. The prediction sets come with frequentist coverage guarantees, meaning that, without any additional assumptions on the original model, the probability of the true value of the computer code at a new point lying within the metamodel prediction interval will be above a chosen confidence threshold. The only key assumption necessary for constructing CP sets is the \emph{exchangeability} of the dataset, which means that the concatenation of the training data set with the new test point are interchangeable in law, which is typically the case when dealing with independent and identically distributed (i.i.d.) samples such as those obtained from a crude Monte Carlo DoE in UQ of computer models or as encountered in many standard ML datasets.

A primary challenge in CP lies in attaining \emph{adaptive} prediction intervals, which refers to the property of varying the interval width for different test points. The concept of ``adaptivity'' \citep{romano_conformalized_2019} is intrinsically tied to the \emph{expressivity} of the metamodel, as the interval width should be small when the metamodel prediction error is minimal and large otherwise. For the purpose of GP quality evaluation therefore, adaptivity of CP interval candidates is a crucial feature.

Three main family of methods exist for building CP intervals: the historical ``full-conformal'' paradigm \citep{vovk_gammerman_shafer_book_2005}, the ``split-conformal'' case and the ``cross-conformal'' setting \citep{gentle_intro_angelopoulos_2023}. For the standard CP estimators in these settings, adaptivity is often lacking and the exploration of non-conformity scores for ensuring this property has been predominantly studied and developed in the split-conformal case \citep{lei_sell_rinaldo_tibshirani_wasserman_2018, romano_conformalized_2019, seedat23_self_supervised}. This approach is not practical in cases with limited budgets and/or database size (which can be the case for costly-to-evaluate models in industrial applications). The split-conformal paradigm necessitates the allocation of a \emph{calibration} set, dividing the available data into three parts for training the metamodel, calibrating the prediction sets, and testing, respectively.
Conversely, the cross-conformal paradigm LaTeX Info: Redefining \BBA on input line 1705.
and especially the ``Jackknife+'' interval estimators \citep{foygel_barber_candes_2021} allows for the utilization of the entire dataset but requires an additional computational budget since it implies training multiple LOO metamodels. 

In this paper, we propose a strategy for obtaining adaptive prediction intervals for a GP in the pure cross-conformal case and use this in the context of GP metamodel quality assessment. We further demonstrate its ability for discriminating between different choices of  prior kernels on a number of numerical examples for reference databases, as well as an industrial use case.

%%-----------------------------%%
\subsection{Related work}
%%-----------------------------%%

Within the full-conformal paradigm, the concept of ``conformalizing'' GPs can be traced back to the \emph{Burnaev-Wasserman program} \citep{vovk_gammerman_shafer_book_2005}. A theorem establishes a theoretical comparison between Bayesian credibility sets and CP sets, assuming the Gaussianity of the original model \citep{vovk_burnaev_14}. This limit theorem provides guarantees that the differences between the upper and lower endpoints of the two intervals follow a zero-mean Gaussian distribution asymptotically. The conclusion drawn is that conformalizing under the Gaussian hypothesis is not asymptotically ``worse'' than standard Bayesian credibility sets. Thorough numerical comparisons with Bayesian credibility sets in various scenarios are performed in \citep{burnaev_nazarov_16}.

The full-conformal paradigm extends to spatial Kriging as well, as demonstrated in \citep{mao_martin_reich_2023}, where CP algorithms are developed for non-Gaussian data by establishing conditions for approximate exchangeability. However, it is important to note that full-conformal methods are computationally expensive, requiring a complete grid search on the output space and can quickly become prohibitive \citep{vovk_gammerman_shafer_book_2005, papadopoulos_2023, foygel_barber_candes_2021}. To enhance efficiency, recent work explores the idea of conformalizing GPs, as discussed in \citep{papadopoulos_2023}.

Moreover, the conformal paradigm finds application in enhancing Bayesian optimization, particularly when GPs serve as query functions \citep{stanton_maddox_wilson_23}. This is particularly relevant when Bayesian credible sets obtained are deemed unreliable due to model misspecification.

An idea for combining the use of a calibration set and the Jackknife+ for obtaining adaptive intervals has been explored in the recent work of \citep{deutschmann_2023}; however pure cross-conformal adaptive methods are not found in recent literature. 

%%-----------------------------%%
\subsection{Contributions and organization}
%%-----------------------------%%

In this work, we introduce a non-conformity score tailored for the use of GP surrogate models within the cross-conformal Jackknife paradigm. Utilizing this score, we derive an adaptive prediction interval named ``J+GP'', along with its min-max variant, and establish their theoretical marginal coverage. The adaptivity of these set-estimators is quantified, and we demonstrate that the length of these intervals is interpretable as a good proxy for surrogate precision. This interpretation is supported by a significant statistical correlation observed between the interval widths and the absolute value of the metamodel error, showcasing their capability to assess the quality of a GP. 

We provide a high-quality implementation of the methodology through a plug-and-play GitHub repository that can be found at the following address: \url{https://github.com/vincentblot28/conformalized\_gp}. This repository is based on two pre-existing Python libraries: OpenTURNS (Open source initiative for the Treatment
of Uncertainties, Risks'N Statistics), an open source UQ platform \citep{openturns_2015}, and MAPIE (Model Agnostic Prediction Interval Estimator), a library employed for CP \citep{mapie_2023}.

The rest of the paper is organized as follows. We recall the definition and main properties of GPs and Conformal Predictors (Section~\ref{sec2}). We then proceed in defining the J+GP conformal predictor estimator and its variants (Section~\ref{secJ+GP}). A methodology for validating the link with the error spread and the adaptivity is then presented. Numerical comparisons between credibility intervals of the GP and the J+GP variants are shown on a selection of databases for regression tasks (Section~\ref{secnumerics}). Finally, Section~\ref{sec:conclusion} draws the main conclusions of this work and discuss a few perspectives.

%%==================================%%
\section{Notations and background}
\label{sec2}
%%==================================%%

In the rest of this paper we suppose fixed a probability space $(\Omega, \mathcal{F}, \mathbb{P})$. Random variables are denoted with capital letters. $\mathcal{N}(\mu,\sigma^2)$ denotes the Gaussian distribution with mean $\mu$ and standard deviation $\sigma$ while $\mathcal{T}(a,b,c)$ denotes the triangular distribution centered in $b\in[a,c]$. $g:\mathcal{X}\to \mathcal{Y}$ denotes a deterministic function where $\mathcal{X}\subset \mathbb{R}^{d}$ and $\mathcal{Y}\subseteq\mathbb{R}$ are regular domains. For a set $D$, $\mathds{1}_{D}$ denotes the indicator function of $D$. The cardinal of the output space will be denoted $n_{grid} = \text{Card}(\mathcal{Y
})$. The Cartesian product of the two spaces is denoted $\mathcal{Z} = \mathcal{X}\times\mathcal{Y}$ and $2^{\mathcal{L}}$ will denote the set of subspaces of a set $\mathcal{L}$. For a given $N\in\mathbb{N}$ we fix an i.i.d. dataset $\mathcal{D}_{N}$ of size $N$ whose elements will be written equivalently as $ Z_{i} = (X_{i}, g(X_{i})) = (X_{i}, Y_{i})$ for all $i\in\{1,\ldots,N\}$. We denote the features $\bm{X} = \{X_{1},\ldots,X_{N}\}$ and similarly the labels $g(\bm{X}) = (g(X_{1}),\ldots,g(X_{N}))$ and the dataset will sometimes be written $\mathcal{D}_{N} = (\bm{X}, g(\bm{X}))$. As customary in supervised ML, the dataset in split into training and testing subsets, $\mathcal{D}_{N} = \mathcal{D}_{n}\cup\mathcal{D}_{m}$ where $N = n +m$ are the respective size of the two subsets. We denote by $\widehat{g}$ a metamodel of $g$ trained on $\mathcal{D}_{n}$ and $\widehat{g}_{-i}$ the corresponding LOO metamodel on $\mathcal{D}_{n}\backslash (X_{i}, g(X_{i}))$ with $i\in\{1,\ldots, n\}$. The Spearman correlation coefficient, corresponding to the Pearson coefficient in the rank space, is denoted by $r_{s}$. The space of continuous $k$-differentiable functions on $\mathcal{L}$ is denoted $\mathcal{C}^{k}(\mathcal{L})$. The space of square $n\times n $ matrices on a set $\mathcal{L}$ will be denoted $\mathcal{M}_{n}(\mathcal{L})$. For an interval $I\subset \mathbb{R}$ we denote its length as $\ell(I)$. For any $m\in \mathbb{N}$, $\mathfrak{S}(m)$ denotes the set of permutations over $\{1,\ldots,m\}$.
For any finite subset $\{v_{i}\}_{i=1,\ldots,n}$ of an ordered set, the $(1-\alpha)$-empirical-quantile, with $\alpha \in (0, 1)$, is given by:
\begin{equation}
\widehat{q}^{\;+}_{n,\alpha}\left\{ v_{i} \right\}:= \text{the}\;\lceil (1-\alpha)(n+1) \rceil \text{-th smallest value of} \; v_{1},\ldots,v_{n}
\end{equation}
with $\lceil \cdot \rceil$ denotes the ceil function. Similarly, the $\alpha$-empirical-quantile is given by:
\begin{equation}
    \widehat{q}^{\;-}_{n,\alpha}\left\{ v_{i} \right\}:= \text{the}\;\lfloor \alpha(n+1) \rfloor \text{-th smallest value of} \; v_{1},\ldots,v_{n} = -\widehat{q}^{\;+}_{n,\alpha}\left\{ - v_{i} \right\}
\end{equation}
where $\lfloor \cdot \rfloor$ denotes the floor function.
%%-----------------------------%%
\subsection{Gaussian Process metamodels}
\label{sec3}
%%-----------------------------%%

%%%%%%%%%%%
\subsubsection{General definitions}
\begin{figure}[!ht]
    \centering
\includegraphics[width=1.0\textwidth]{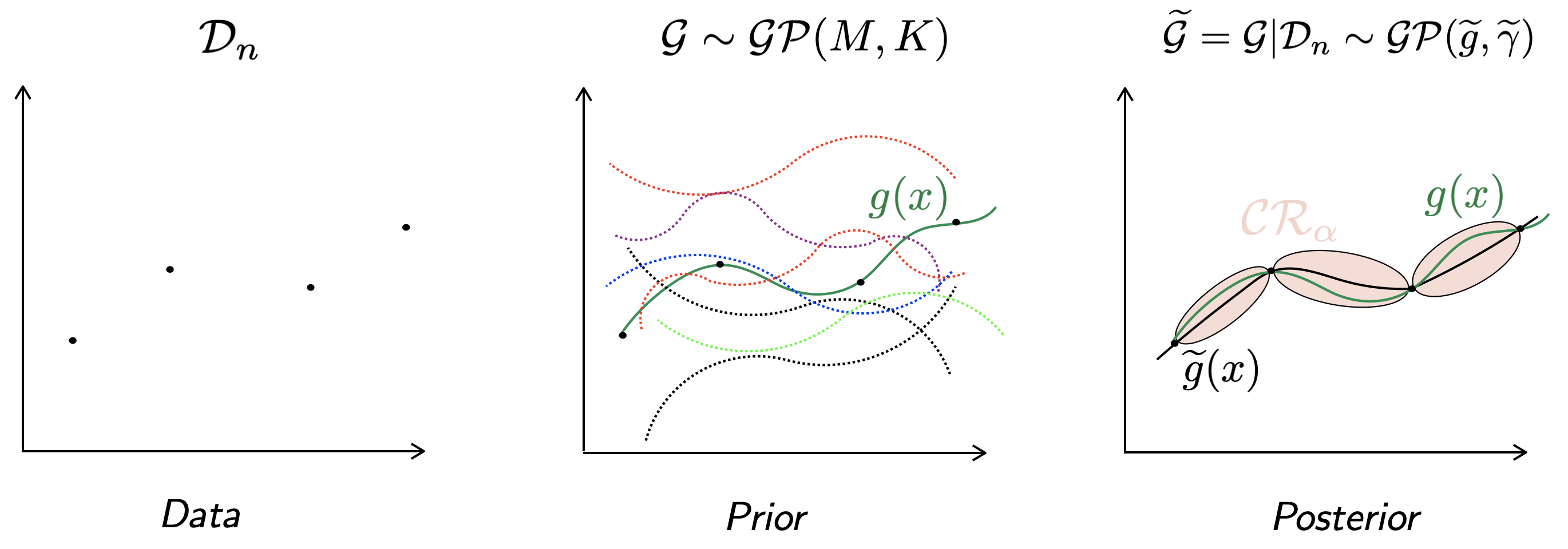}
    \caption{GP regression metamodeling illustration. The data obtained from the numerical code is modeled with a prior GP and is then conditioned by the data. In the absence of noise, the posterior process interpolates the data and produces \emph{credibility} intervals.}
    \label{fig:GP}
\end{figure}

%%%%%%%%%%%
To build a GP surrogate model \citep{rasmussen_williams_book_2006}  of function $g$, we start by supposing that $g$ is the realization of a certain process $\mathcal{G} = \mathcal{GP}(M,K)$, where $M:\mathcal{X}\to \mathcal{Y}$ is the \emph{mean} of the process and $K:\mathcal{X}\times\mathcal{X}\to \mathbb{R}$ is the \emph{covariance kernel} of the process. Then, this process is \emph{conditioned} on the available dataset $\mathcal{D}_{n}$. By doing so, this procedure is simply a Bayesian regression method while considering a Gaussian \emph{prior} on the function $g$ in order to then obtain a \emph{posterior} distribution. We sketch the general principle of this type of regression in Figure~\ref{fig:GP}. For simplicity we choose that $M=0$ (corresponding to the usual case of ``ordinary Kriging'') and we use Matérn-$\nu$ kernels defined for all $\nu = (2k + 1) / 2,\; k \in \mathbb{N}$ and $x,x'\in \mathcal{X}$ by:
\begin{equation}
\begin{aligned}
K(x, x') &= K_{(\nu,\theta,\sigma)}(x, x') \\
    &= \sigma^2\frac{2^{1-\nu}}{\Gamma(\nu)}\left( \sqrt{2\nu}\frac{\lvert x - x'\rvert}{\theta}\right)^{\nu}K_{\nu}\left(\sqrt{2\nu}\frac{\lvert x - x'\rvert}{\theta}\right).
\end{aligned}
\end{equation}
Here $K_{\nu}$ is a Bessel function of the second-type and $\Gamma$ is the Euler gamma function. This kernel allows to better control the regularity of the process through its hyperparameter $\nu$ since the corresponding sample paths will lie in $\mathcal{C}^{\lfloor \nu - 1\rfloor}(\mathcal{X})$ \citep{gu_wang_berger_2018}. The final conditional process $\widetilde{\mathcal{G}}:= \mathcal{G}|\mathcal{D}_{n}$ is a GP with posterior mean and covariance functions defined for all $x,x' \in \mathcal{X}$ as:
\begin{equation}
    \widetilde{g}(x):= k(x)^{T}\bm{K}^{-1}g(\bm{X}),\;\text{and}\;\; \widetilde{K}(x,x'):= K(x,x') - k(x)^{T}\bm{K}^{-1}k(x'),
\end{equation}
where for all $x\in\mathcal{X}$:
\begin{equation}
    k(x):=(K(x,X_{1}),\ldots,K(x,X_{n}))^{T}\in\mathbb{R}^n,\;\text{and}\;\bm{K}:= (K(X_{i},X_{j}))_{1\leq i,j \leq n}\in\mathcal{M}_{n}(\mathbb{R}).
\end{equation}

The mean of the posterior process $\widetilde{g}$ will act as a metamodel for the deterministic function $g$, thus in the case of GPs, $\widehat{g} = \widetilde{g}$.
For a choice of the regularity parameter $\nu$, the hyperparameters $(\sigma^{2},\theta)$ of the Matérn kernel can be optimized either through a maximum likelihood estimation (MLE) or through cross-validation (CV) \citep{acharki_robust_2023}. In the case of ordinary Kriging, the MLE problem is:
\begin{equation}
    (\sigma_{\textnormal{MLE}}^{2}, \theta_{\textnormal{MLE}}) \in \argmin_{(\sigma^{2}, \theta)} \left(g(\bm{X})^{T}\bm{K}^{-1}g(\bm{X}) + \log(\det \bm{K}) \right).
\end{equation}
The MLE procedure yields better results if the kernel type is well-specified; while the CV method is more robust in case of misspecification \citep{acharki_robust_2023}.
For $x\in\mathcal{X}$, the posterior standard deviation is denoted by:
\begin{equation}
    \widetilde{\gamma}(x):= \widetilde{K}^{1/2}(x,x).
\end{equation}
In this setting, it can be shown that for $x = X_{1}, \ldots, X_{n}$ then $\widetilde{\gamma}(x) = 0$ and $\widetilde{g}(x) = g(x)$ and thus, the GP metamodel is interpolating. In the case where the results of the code are perturbed with noise such that:
\begin{equation}
    Y_{i} = g(X_{i}) + \epsilon,
\end{equation}
then the covariance matrix has to take into account a so-called \emph{nugget effect}, meaning that one needs to add a regularization term $\sigma_{\epsilon} I_{n}$ such that:
\begin{equation}
    \bm{K}_{\epsilon}:= \bm{K} + \sigma_{\epsilon} I_{n},
\end{equation}
modeling the noise dispersion. The new hyperparameter $\sigma_{\epsilon}$ has to be tuned by constructing a full-likelihood and the resulting metamodel will no longer be interpolating \citep{iooss_marrel_2017}.
%This type of modeling is not addressed in the present work but will be discussed as a perspective for future work.

Moreover a ``Full-Bayesian'' approach exists for obtaining the posterior distribution of the hyperparameters and updating the predictive distribution. This method is out of the scope of this work and is likely infeasible for the purpose of constructing cross-conformal predictors since it would involve tuning complex Monte Carlo Markov chain algorithms on too many LOO metamodels.

For our purposes, we will be working with MLE-optimized hyperparameters of Matérn-$\nu$ kernels in the ordinary Kriging scenario (corresponding to zero prior mean) and estimating empirically the nugget effect.

%%%%%%%%%%%
\subsubsection{Bayesian credibility intervals}
%%%%%%%%%%%
In Bayesian inference, a credibility interval is related to the distribution of a parameter of the posterior law. For a certain credibility level $1 - \alpha \in (0,1)$, the true value of the parameter would lie in this interval with probability $1 - \alpha$ given the available data. In the case of GPs, the parameter is the mean of the posterior GP and for any new point $X_{n+1}\in\mathcal{X}\backslash\bm{X}$, the credibility prediction interval is given by:
\begin{equation}
     \mathcal{CR}_{\alpha}(X_{n+1}) = \left[ \;\widetilde{g}(X_{n+1}) \pm u_{1-\alpha/2}\widetilde{\gamma}(X_{n+1}) \;\right],
\end{equation}
where $u_{1-\alpha/2}$ is the $(1-\alpha/2)$-quantile of the standard Gaussian distribution. Under the Gaussian assumption on the original function, if $g$ was \emph{truly} modeled by our posterior $\mathcal{G}|\mathcal{D}_{n}$, then we would have the conditional coverage:
\begin{equation}
    \mathbb{P}\left( g(X_{n+1}) \in \mathcal{CR}_{\alpha}(X_{n+1}) \;| \;\mathcal{D}_n\right) = 1- \alpha,
\end{equation}
where $\mathcal{CR}_{\alpha}(X_{n+1})$ would be a \emph{prediction} interval for the true function $g$. In practice we do not have access to the distribution $Q$ of the new point $(X_{n+1},g(X_{n+1}))$. Therefore there is not guarantee to have:
\begin{equation}
    Q(g(X_{n+1}) \in \mathcal{CR}_{\alpha}(X_{n+1})) = 1- \alpha.
\end{equation}

Even if the Gaussian assumption holds true, a common occurrence is the \emph{misspecification} of the posterior model. This implies that the set of prior mean and/or the family of kernels is proven to be incorrect. In essence, the practical reliability of Bayesian credibility intervals, particularly for GPs, can be significantly compromised.

%%-----------------------------%%
\subsection{Conformal Prediction Intervals}
%%-----------------------------%%

%%%%%%%%%%%
\subsubsection{General definition}
%%%%%%%%%%%
CP is a finite-sample and distribution-free framework for building prediction sets with a statistical guarantee on the coverage rate for any predictive algorithm. Suppose a given dataset $\mathcal{D}_n$ and a new test point $Z_{n+1} = (X_{n+1}, Y_{n+1})$. It is assumed that the $n+1$ points are exchangeable \citep{vovk_gammerman_shafer_book_2005}.
Formally this means that for any permutation $\pi \in \mathfrak{S}(n+1)$, we have:
\begin{equation}
   (Z_{1}, \ldots, Z_{n+1}) \overset{\mathcal{L}}{=} (Z_{\pi(1)}, \ldots, Z_{\pi(n+1)}).
\end{equation}
More concretely, this means that $Z_{n+1}$ could have been used as a training point and that any training point could have been a test point. An i.i.d. dataset is a special case of an exchangeable dataset. For any confidence level $\alpha\in (0,1)$, a conformal predictor of coverage $1-\alpha$ is any measurable function of the form \citep{vovk_gammerman_shafer_book_2005}:
\begin{equation}
\begin{aligned}
        C_{\alpha} \colon \mathcal{Z}^{n}\times \mathcal{X} & \to 2^{\mathcal{Y}}\\
        (\mathcal{D}_{n}, X) & \mapsto C_{n,\alpha}(X),
\end{aligned}
\end{equation}
such that for a new test point $Z_{n+1}$:
\begin{equation}
    \mathbb{P}\left(Y_{n+1} \in C_{n,\alpha}(X_{n+1})\right) \geq 1 - \alpha.
    \label{eq:coverage}
\end{equation}

To build estimators of such set-functions, one relies on the use of a non-conformity score. 
This score defined as a measurable function of the form \citep{vovk_gammerman_shafer_book_2005}:
\begin{equation}
    \begin{aligned}
            R \colon \mathcal{Z}^{n}\times \mathcal{Z} & \to \mathbb{R}\\
        (\mathcal{D}_{n}, Z) & \mapsto R(\mathcal{D}_{n}, Z),
\end{aligned}
\end{equation}
that quantifies how ``strange'' is the point $Z$ compared to the dataset $\mathcal{D}_{n}$. For example, if a metamodel $\widehat{g}$ of a code $g$ has been trained on $\mathcal{D}_{n}$ then a straightforward non-conformity score is given by the residuals:
\begin{equation}
    R(\mathcal{D}_{n}, Z_{n+1}) = \lvert g(X_{n+1}) - \widehat{g}(X_{n+1})\rvert.
    \label{eq:non-conf}
\end{equation}
It is noteworthy to insist that in practice the coverage property~\eqref{eq:coverage} is \emph{marginal}, by that meaning that it is averaged over all possible realizations of the training set $\mathcal{D}_{n}$. A more standard coverage is the \emph{training-conditional} coverage property \citep{gentle_intro_angelopoulos_2023}, meaning that for a conformal predictor estimator $\widehat{C}_{n,\alpha}$, one has:
\begin{equation}
    \mathbb{P}\left(Y_{n+1} \in \widehat{C}_{n,\alpha}(X_{n+1})\;|\;\mathcal{D}_{n}\right) \geq 1-\alpha.
\end{equation}
There are mainly three ways of estimating such conformal predictors: full-CP \citep{vovk_gammerman_shafer_book_2005} (also called ``transductive CP''), split-CP \citep{papadopoulos_induc1, papadopoulos_induc2} (also called ``inductive CP''), and cross-CP \citep{vovk:15}.

Transductive CP is historically the first CP method introduced by \citep{vovk_gammerman_shafer_book_2005}. For any choice of a non-conformity score, it implies computing the following set:
\begin{equation}
    \widehat{C}_{n,\alpha}(X_{n+1}) = \left\{Y\in \mathcal{Y}:\;\frac{\text{Card}(\{i\in\{1,\ldots,n\}, R(
\widetilde{\mathcal{D}}_{n}, Z_{i}) \geq R(\widetilde{\mathcal{D}}_{n},(X_{n+1}, Y))\})}{n}\right\},
\end{equation}
where $\widetilde{\mathcal{D}}_{n} = \mathcal{D}_{n}\cup \{Z_{n+1}\}$. As mentioned in the introduction, transductive CP is computationally intensive as it involves the training of one metamodel for each possible value in $\mathcal{Y}$ as can be seen if the regular residual non-conformity score~\eqref{eq:non-conf} is used. This conformal predictor can be made computationally effective in the cases of Ridge and Lasso regressions \citep{lei_lasso, vovk_ridge_2001}, $k$-Nsearest neighbors algorithm \citep{papdopoulos_knn_2008, papadopoulos_knn_2011} and more recently, GP regression \citep{papadopoulos_2023}. 

Inductive -- or split-CP -- on the other hand, has a very low computational cost as it requires a single training of the learning model. However, it needs a calibration (or holdout) set which contains observations which have not been used during the training phase. This set is then used for estimating the quantiles of the evaluated non-conformity scores (usually, the residuals) on this set for constructing the intervals. When only a few hundred observations are available, such a calibration set may be impossible to obtain. 

Unlike the first two techniques, cross-CP has a relatively low computational cost and does not necessitate any holdout set. 
We now proceed in presenting the cross-conformal predictors.

%%%%%%%%%%%
\subsubsection{Cross-conformal prediction sets}
%%%%%%%%%%%
The standard Jackknife prediction intervals require learning a metamodel $\widehat{g}$ on a training dataset $\mathcal{D}_{n}$ and also $n$ leave-one-out (LOO) metamodels $\widehat{g}_{-i}$, with $1 \leq i \leq n$. It then makes use of the $(1-\alpha)$-empirical quantile of the LOO residuals defined by:
\begin{equation}
    R_{i}^{\textnormal{LOO}}:= \lvert g(X_{i}) - \widehat{g}_{-i}(X_{i}) \rvert.
\end{equation}
For a new point $X_{n+1}$ and a coverage level $1- \alpha$, the standard Jackknife interval is defined by: 
\begin{equation}
\widehat{C}_{n,\alpha}^{J}(X_{n+1}) = \left[\widehat{g}(X_{n+1}) \pm \widehat{q}_{n,\alpha}^{\;\pm}\left\{R_{i}^{\textnormal{LOO}}\right\}\right].
\end{equation}

Unfortunately this prediction interval does not fulfill the coverage property~\eqref{eq:coverage}, especially in the case of a small dataset as mentioned in \citep{foygel_barber_candes_2021}. To circumvent this limitation, a more robust cross-conformal estimator is the Jackknife+ \citep{foygel_barber_candes_2021}. In this case, the interval is not centered on the prediction of the fully-trained metamodel but the LOO predictions are added in the empirical quantile. The estimator is given by:
\begin{equation}
    \widehat{C}^{J+}_{n,\alpha}(X_{n+1}) = \left[\widehat{q}^{\;\pm}_{n,\alpha}\left\{\widehat{g}_{-i}(X_{n+1}) \pm R_{i}^{\textnormal{LOO}}\right\}\right].
\end{equation}
This estimator has the universal coverage property of $1-2\alpha$. As mentioned in Theorem $5$ of \citep{foygel_barber_candes_2021}, the factor ``2'' in $1-2\alpha$ can be removed if the metamodel satisfies the out-sample stability i.e., for any $\epsilon > 0$, one can find $\lambda\in [0,1]$ such that for all $i\in\{1,\ldots,n\}$:
\begin{equation}
    \label{eq:out_sample}
    \mathbb{P}(|\widehat{g}(X_{i}) - \widehat{g}_{-i}(X_{i})| \leq \epsilon) \geq 1-\lambda.
\end{equation}
The authors of the present paper do not know whether such stability exists for GP metamodels.
The authors in \citep{foygel_barber_candes_2021} indicate that, in numerical applications, the empirical coverage barely drops below $1-\alpha$ unless the case-study is somewhat ``pathological''. More recently, it has been established that the training-conditional property~\eqref{eq:traning_cond} is achieved under the out-sample stability~\citep{liang_stability_2023}. In most empirical cases however, the coverage property is respected.

\begin{figure}%[!ht]
    \centering
\includegraphics[width=1.0\textwidth]{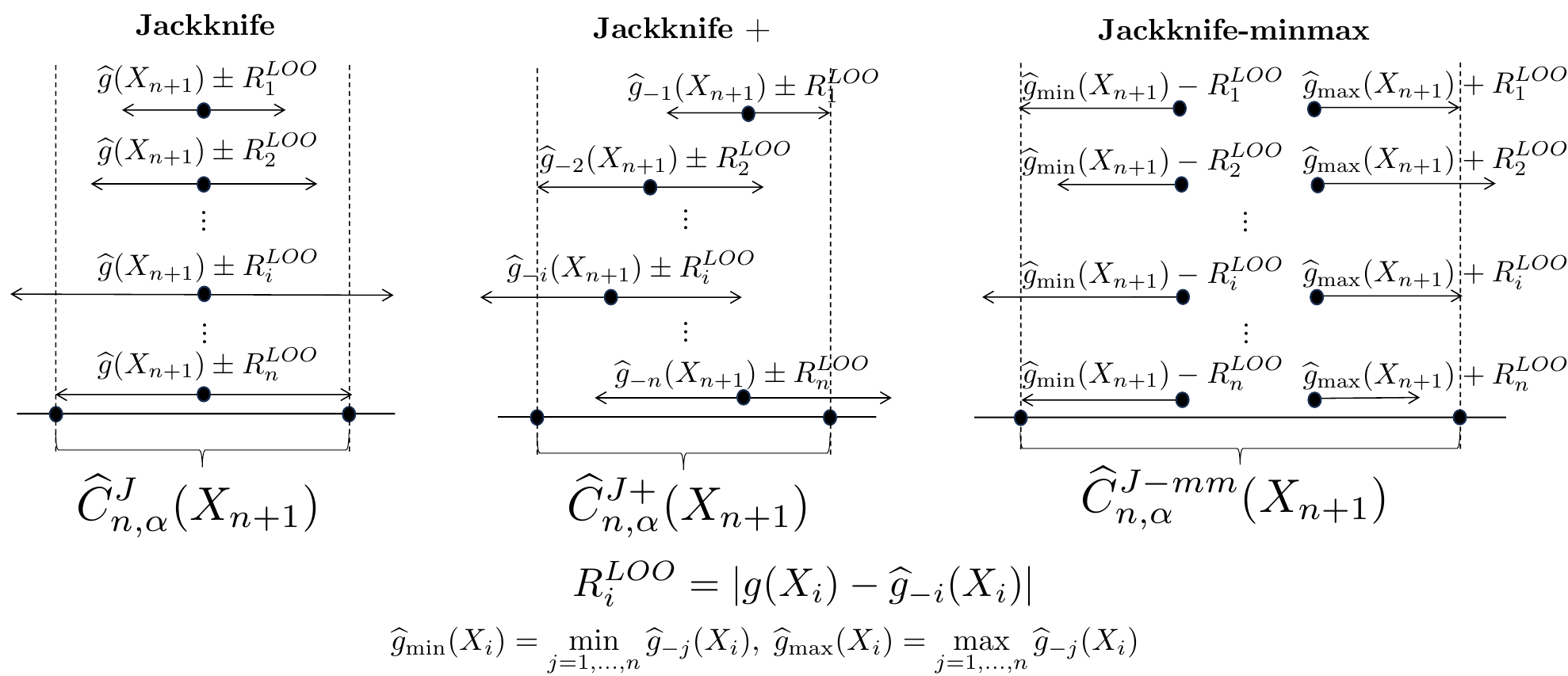}
    \caption{Illustration of the different cross-CP methods.}
    \label{fig:plus_minmax}
\end{figure}

Another way of achieving the $1-\alpha$ coverage is through the ``Jackknife-minmax'' method \citep{foygel_barber_candes_2021} which is a more conservative implementation of the Jackknife+ method. This method differs from the latter as it does not use the prediction of each LOO model but the minimum (resp., the maximum) predicted value to compute the lower (resp., the upper) confidence bound. The Jackknife-minmax estimator is given by the following expression: 
\begin{equation}
\begin{split}
    \widehat{C}^{\textnormal{J-mm}}_{n,\alpha}(X_{n+1}) = \left[
    \min_{i=1, ..., n} \left\{\widehat{g}_{-i}(X_{n+1})\right\} - \widehat{q}^{\;-}_{n,\alpha}\left\{R_{i}^{\textnormal{LOO}}\right\},\right. \\
    \left. \max_{i=1, ..., n} \left\{\widehat{g}_{-i}(X_{n+1})\right\} + \widehat{q}^{\;+}_{n,\alpha}\left\{R_{i}^{\textnormal{LOO}}\right\} 
    \right].
\end{split}
\end{equation}
An illustration adapted from \citep{foygel_barber_candes_2021} of the different Jackknife methods can be found in Figure~\ref{fig:plus_minmax}.
Even if the Jackknife methods have a lower computational cost than the full-CP, if one disposes of a dataset with more than a few thousand observations, or in the case where the metamodel $\widehat{g}$ is very long to train, these cross-CP methods can still be too expensive. A recap of the different cross-CP methods mentioned with their coverage and computational cost can be found in Table~\ref{table:recap_cp}. One drawback of this CP method is that there is no theoretical guarantee for adaptivity. This is what will be addressed in the following with the help of GPs.

\begin{table}[hbtp]
  \caption{Theoretical coverage and reminder of both training and evaluation costs for CP methods \citep{foygel_barber_candes_2021}.}
  \label{table:recap_cp}
  \begin{tabular}{rccc}
  \toprule
  \bfseries Method & \bfseries Theoretical coverage & \bfseries Training cost & \bfseries Evaluation cost \\
  \midrule
  Full & $\geq 1-\alpha$ & $m \cdot n_{grid}$ & $n \cdot m \cdot n_{grid}$ \\
  Split & $\geq 1-\alpha$ & $1$ & $m$ \\
  Jackknife+ & $\geq 1-2\alpha$ & $n$ & $n \cdot m$ \\
  Jackknife-minmax & $\geq 1-\alpha$ & $n$ & $n \cdot m$ \\
  \bottomrule
\end{tabular}  
\end{table}

%%==================================%%
\section{Conformalized GPs: the J+GP method}
\label{secJ+GP}
%%==================================%%

%%-----------------------------%%
\subsection{Motivations and estimator}
%%-----------------------------%%

The idea is to adapt the Jackknife+ method to GP metamodels for obtaining adaptive prediction intervals. Suppose we have conditioned the Gaussian surrogate model on a dataset $\mathcal{D}_{n}$ by optimizing the hyperparameters $(\sigma^{2}_{\textnormal{MLE}}, \theta_{\textnormal{MLE}})$ of a Matérn-$\nu$ kernel with given $\nu$. We thus have access to the posterior mean $\widetilde{g}$ as well as the posterior standard deviation denoted by $\widetilde{\gamma}$. We will also make use of different integer-powers of the posterior variance as proposed in \citep{papadopoulos_2023}. For an integer $\beta$, we will write $\widetilde{\gamma}^{\beta}(.) = \widetilde{K}(.,.)^{\beta/2}$. For the respective LOO posteriors, we write $\widetilde{g}_{-i}$ and $\widetilde{\gamma}_{-i}$ for all $i\in\{1,\ldots,n\}$. We proceed in defining the LOO Gaussian non-conformity score, fixing a small $\delta > 0$ and an integer root $\beta$:
\begin{equation}
    R_{i}^{\textnormal{LOO}\gamma}:= \frac{\lvert g(X_{i}) - \widetilde{g}_{-i}(X_{i}) \rvert}{\max\left(\delta, \widetilde{\gamma}^{\;\beta}_{-i}(X_{i})\right)},\;\;\forall i\in\{1,\ldots,n\}.
\end{equation}
The interest of $\delta$ is to avoid the zero division when the metamodel is interpolating (e.g., in absence of a nugget effect). For a new prediction point $X_{n+1}\in\mathcal{X}$ and a coverage rate $1 - \alpha \in (0,1)$, we define the ``J+GP'' conformal predictors which are a variant of the Jackknife+ estimator adapted to the GP metamodeling setting:
\begin{equation}
    \widehat{C}^{\textnormal{J+GP}}_{n,\alpha}(X_{n+1}) = \left[ \widehat{q}^{\;\pm}_{n,\alpha}\left\{ \widetilde{g}_{-i}(X_{n+1}) \pm R_{i}^{\textnormal{LOO}\gamma} \times \max\left(\delta, \widetilde{\gamma}^{\;\beta}_{-i}(X_{n+1})\right)\right\}\right]. 
\end{equation}
This estimator enables adaptivity at different prediction points since the edges of the intervals are controlled by a function of $X_{n+1}$. Moreover, for this estimator, we achieve the same theoretical coverage as the Jackknife+ as presented in the following theorem.

\smallskip
\begin{theorem}{}
Assume $\mathcal{D}_{n}$ is exchangeable. For a new point $X_{n+1}\in\mathcal{X}$ and a coverage level $1 - \alpha\in(0,1)$:
\begin{equation}
\mathbb{P}\left(g(X_{n+1})\in \widehat{C}^{\textnormal{J+GP}}_{n,\alpha}(X_{n+1})\right) \geq 1 - 2\alpha.
\end{equation}
\label{th:theorem_1}
\end{theorem}

\smallskip
The proof is provided in Appendix~\ref{secA1} and is based \emph{mutatis mutandis} on the proof for the Jackknife+ in \citep{foygel_barber_candes_2021}. We similarly propose the following ``J-minmax-GP'' estimator: 
\begin{equation}
\begin{split}
    \widehat{C}^{\textnormal{J-mm-GP}}_{n,\alpha}(X_{n+1}) = \left[
    \min_{i=1, ..., n} \widetilde{g}_{-i} - \widehat{q}^{\;-}_{n,\alpha}\left\{R_{i}^{\textnormal{LOO}\gamma}\times \max\left(\delta, \widetilde{\gamma}^{\;\beta}_{-i}(X_{n+1})\right)\right\}\right.,\\ 
    \left.\max_{i=1, ..., n} \widetilde{g}_{-i} + \widehat{q}^{\;+}_{n,\alpha}\left\{R_{i}^{\textnormal{LOO}}\times \max\left(\delta, \widetilde{\gamma}^{\;\beta}_{-i}(X_{n+1})\right)\right\} 
    \right],
\end{split}
\end{equation}
that inherits the same coverage guarantee as the standard min-max estimator, as shown by the following theorem.

\smallskip
\begin{theorem}{}
Assume $\mathcal{D}_{n}$ is exchangeable. For a new point $X_{n+1}\in\mathcal{X}$ and a coverage level $\alpha\in(0,1)$:
\begin{equation}
\mathbb{P}\left(g(X_{n+1})\in \widehat{C}^{\textnormal{J-mm-GP}}_{n,\alpha}(X_{n+1})\right) \geq 1 - \alpha.
\end{equation}
\label{th:theorem_2}
\end{theorem}

\smallskip
The proof of the preceding theorem is  found in the Appendix~\ref{secA2} and is similar to the proof for the Jackknife-minmax in \citep{foygel_barber_candes_2021}.

%%-----------------------------%%
\subsection{Methodology evaluation}
%%-----------------------------%%

To assess the capabilities of the J+GP and J-minmax-GP estimators in comparison with the classical cross-CP and the Bayesian credibility sets, a two-step approach is considered. In the following, let $\widehat{C}^{*}_{n,\alpha}$ denote any type of prediction interval. The following computations are performed on the test subset $\mathcal{D}_{m}$.
We check whether the empirical-coverage property is achieved for different values of the $1-\alpha$ coverage rate:
\begin{equation}
    \frac{1}{m}\sum_{i=1}^{m}\mathds{1}\left\{g(X_{i})\in\widehat{C}^{*}_{n,\alpha}(X_{i})\right\}  \approx 1-\alpha, \;\;\forall \;0\leq \alpha \leq 1.
    \label{eq:empirical_cov}
\end{equation}
Secondly, the correlation of the interval width and the model error is computed. Indeed, for the intervals to be informative, they have to be small when the prediction error is small and large otherwise. Therefore we could expect a significant correlation between the two. This metric will quantitatively reflect the adaptive nature of our intervals. It is thus valid to verify that, for a given coverage $1-\alpha\in(0,1)$, the Spearman correlation on the test database is non-null with statistical significance, i.e.:
\begin{equation}  
    0 \ll r_{s}\left(\{(\ell(\widehat{C}^{*}_{n,\alpha}(X_{i})), \lvert g(X_{i}) - \widetilde{g}(X_{i})\rvert) \}_{i\in\{1,\ldots,m\}}\right)
    \label{eq:empirical_corr}
\end{equation}
Here the Spearman correlation is chosen for avoiding the interpretation to be falsified by outliers. For achieving statistical significance, we compute bootstrap intervals on the estimation of the correlation metric.

Concerning the quality of the metamodel, it is assessed with the help of the predictivity coefficient (see, e.g., \citep{fekioo23}):
\begin{equation}
    Q^{2} = 1 - \sum_{i=1}^{m} \frac{\left| g(X_{i}) - \widetilde{g}(X_{i})\right|^{2}}{\text{Var}({g(X_{i})})}.
\label{eq:q2_form}
\end{equation}

Usually, this is the main index computed for assessing the predictive power of the surrogate model and to ensure its validation. The closer the $Q^{2}$ is to $1$, the more predictive the metamodel is. Here, the analysis can be completed with the computation of the empirical coverage rates and of the correlations between the lengths of the intervals and the residuals. Moreover, this strategy provides a pathway for a decision-maker to assess which model suits his application best, since one can deploy it with various priors on the covariane kernel and the mean in the scope of further enhancing the predictive power for the final metamodel. In our numerical examples, we demonstrate this by-product of our methodology by testing several GPs with different values of  Matérn regularity parameter $\nu$ and show that it allows to discriminate between them in order to choose the best one. Here, an emphasis is put on Matérn kernels since they are widely used in practice. However this approach can also be deployed on other families of kernels.

%%==================================%%
\section{Numerical results}
\label{secnumerics}
%%==================================%%

In order to test the proposed methodology, a series of numerical toy and use cases are carried out. We choose standard regression benchmarks, both from the standard ML and UQ literature, as well as a real engineering use case from nuclear engineering provided by EDF, the French national electric utility company.

In what follows, the ML regression datasets are not generated from a computer code $g$ on a certain DoE but come from real-life observations. For these benchmarks, the contribution of noise becomes non-negligible. This noise is thus taken into account in the GP prior kernel through a nugget effect. As explained in Section~\ref{sec1}, this nugget effect could be optimized (see, e.g., \citep{iooss_marrel_2017}). In this work, we empirically  estimate the noise  and add it by hand, leaving this nugget optimization aspect for future work. As a preliminary step, standardization of the input data is recommended. For cases for which we have access to the input distributions, we rely on the means and standard deviations of these distributions, while for cases for which we only hold a sample of data, we turn to empirical counterparts of these statistics.

In the case of analytical functions, artificial Gaussian noise is added with prescribed standard deviations. For the deterministic codes (as for the EDF database), the noise is assumed to be null and the error corresponds to the metamodel approximation error. 

We start by going over the characteristics of each database and presenting the performance of each GP by computing its predictivity coefficient as recalled in~\eqref{eq:q2_form} and the mean-squared error (MSE). Afterwards we proceed with a detailed study of three databases namely the Computer Hardware (CPU) ML regression task \citep{misc_computer_hardware_29}, the Morokoff \& Caflisch function with noise (analytical UQ benchmark) \citep{morokoff_caflisch_1995} and the EDF industrial use case (named ``TPD'' for ``THYC-Puffer-DEPOTHYC'', see \citep{jaber_et_al_2023}). For each database, we present the performances of the different prediction intervals, namely the GP credibility intervals, the regular Jackknife+ and Jackknife-minmax, and finally our J+GP and J-minmax-GP estimators. Three indices based on the prediction intervals are computed on the test database: \begin{inparaenum}[(a)] \item the empirical coverage rate given in~\eqref{eq:empirical_cov}, \item the average width observed, and \item the Spearman correlation between the width and the error of the metamodel given in~\eqref{eq:empirical_corr}, all these three metrics being computed for three different target-coverage levels ($90\%$, $95\%$ and $99\%$) and three different GP Matérn kernels (namely $\nu = 1/2, 3/2, 5/2$)\end{inparaenum}. In order to have a more robust estimation of the Spearman correlation, we perform a bootstrap estimation with 999 samples. 

For the purpose of model-selection, a threshold has to be established in order to validate the empirical coverage guarantee. One cannot expect to use a hard threshold on the desired coverage rate as there can be some intrinsic fluctuations around this value. To have a soft threshold, denoted by $t$, we use the result proven in \citep{vovk_conditional}, where it is shown that the distribution of the training-conditional coverage has the following analytical form: 
\begin{equation}
    \label{eq:traning_cond}
    t:= \mathbb{P}\left(Y_{n+1}\in \widehat{C}^{*}_{n,\alpha}(X_{n+1})\; | \;\mathcal{D}_{n}\right) \sim \text{Beta}(n+1-l, l),\;\text{with}\; l= \lfloor (n+1)\alpha \rfloor.
\end{equation}
This formula is valid under the training-conditional guarantee while standard cross-CP estimators ensure the coverage over all possible sets of training data and test points (what is called \emph{marginal} coverage). Hence, even if it has not been proved that the GPs achieve the out-sample stability required for guaranteeing the training-conditional coverage \citep{liang_stability_2023}, comparing our empirical coverage to the $\upsilon$-quantile of the $\text{Beta}(n+1-l, l)$ distribution seems to be a reasonable proxy. In the following, we have chosen $\upsilon=0.1$ meaning that the threshold $t$ (above which we consider that coverage is exceeded) is defined such that:
\begin{equation}
    \mathbb{P}(t \geq 1-\alpha ) \geq 1-\upsilon, 
\end{equation}
where $1-\alpha\in(0,1)$ is the desired coverage which follows~\eqref{eq:traning_cond}. Table \ref{tab:cov_th} provides a summary of all the thresholds used for GP selection on the various datasets while considering a panel of confidence levels.

\begin{table}[htbp]
\centering
  \caption{Thresholds above which it is established that the coverage rate is achieved for different confidence levels.}
  \setlength{\tabcolsep}{7pt}
    \begin{tabular}{cccc}
    \toprule
    Dataset & 90\% & 95\% & 99\% \\
    \midrule
    CPU & 0.875 & 0.931 & 0.986 \\
    Morokoff \& Caflisch & 0.882 & 0.938 & 0.985 \\
    TPD & 0.886 & 0.940 & 0.985 \\
    \bottomrule
    \end{tabular}

  \label{tab:cov_th}%
\end{table}

In the rest of this section, we highlight, for every empirical coverage rate above the imposed threshold $t$, the kernel whose GP model has the smallest average width and the surrogate with the highest Spearman correlation (i.e., correlation between the width of the interval and the approximation error). In general, it is not the same kernel that performs best on both metrics. In this case, the decision-maker has to choose, depending on the targeted application, between more expressivity or more conservatism.  

%%-----------------------------%%
\subsection{Code description and availability}
%%-----------------------------%%

Our results have been obtained with a Python code built with the help of the OpenTURNS \citep{openturns_2015} and MAPIE \citep{mapie_2023} open source libraries. A wrapper around OpenTURNS has been implemented to make the Scikit-Learn GP constructors~\citep{pedregosa:11} compatible with it (i.e., with a \textsf{fit} and a \textsf{predict} methods) since MAPIE handles such Scikit-Learn objects. Only few changes have be made to the MAPIE library to make it compatible with our methodology and it preserves all of its standard conformal methods. The corresponding GitHub repository can be found at the following address: \url{https://github.com/vincentblot28/conformalized\_gp}.

%%-----------------------------%%
\subsection{Benchmark on ML regression datasets}
%%-----------------------------%%

%%%%%%%%%%%
\subsubsection{Performance of the trained GPs}
%%%%%%%%%%%

In Table~\ref{tab:q2_mse}, we present the different database sizes, the percentage used for training and testing as well as the empirical value of the nugget effect. For the three different Matérn regularity parameters, the corresponding predictivity coefficient and mean-squared-error are displayed. 

\begin{table}[htbp]
\centering
  \caption{Summary of the performance metrics of the GP metamodels.}
  \setlength{\tabcolsep}{7pt}
    \begin{tabular}{cc|cccc}
\toprule
Matérn &  & CPU & Morokoff\&Caflisch & TPD\\
\midrule
\shortstack{$\;$ \\ $\;$ } & \shortstack{$d$ \\ $N$ \\ $\%\text{train}$ \\ $\%\text{test}$ \\
$\sigma_{\epsilon}$} & \shortstack{$7$   \\ $209$\\ $80$ \\ $20$ \\ $0.1$} & \shortstack{$10$ \\ $600$\\ $75$ \\ $25$ \\ $10^{-4}$} & \shortstack{$7$ \\ $1000$\\ $80$ \\ $20$ \\ $0$} \\
\hline
\shortstack{$1/2$ \\ $\;$ } & \shortstack{$Q^{2}$ \\ MSE} & \shortstack{$0.845$ \\ $8.3\times 10^{3}$} & \shortstack{$0.923$ \\ $2.33\times10^{-3}$} & \shortstack{$0.990$ \\ $1.46$} \\
\hline
\shortstack{$3/2$ \\ $\;$} & \shortstack{$Q^{2}$ \\ MSE} & \shortstack{$0.856$ \\ $7.7\times 10^{3}$} & \shortstack{$0.935$ \\ $1.98\times10^{-3}$} & \shortstack{$0.996$ \\ $0.54$} \\
\hline
\shortstack{$5/2$ \\ $\;$} & \shortstack{$Q^{2}$ \\ MSE} & \shortstack{$0.854$ \\ $7.8\times 10^{3}$} & \shortstack{$0.937$ \\ $1.93\times10^{-3}$} & \shortstack{$0.997$ \\ $0.46$} \\
\bottomrule
\end{tabular}

  \label{tab:q2_mse}%
\end{table}

%%%%%%%%%%%
\subsubsection{CPU Dataset}
%%%%%%%%%%%
This regression database involves predicting CPU performance with $7$ features and a dataset composed of $N=209$ instances \citep{misc_computer_hardware_29}, featuring integer values. Employing an $80\%-20\%$ split between training and testing, all the GPs consistently exhibit strong performance, with a predictivity coefficient showing only slight variation for all the kernels regularity parameter values $\nu$ as can be seen in Table~\ref{tab:q2_mse}. This scenario underscores the potential interest for decision-makers of exploring different indices, related to  our prediction intervals.

\begin{figure}[!ht]
    \centering
\includegraphics[width=0.9\textwidth]{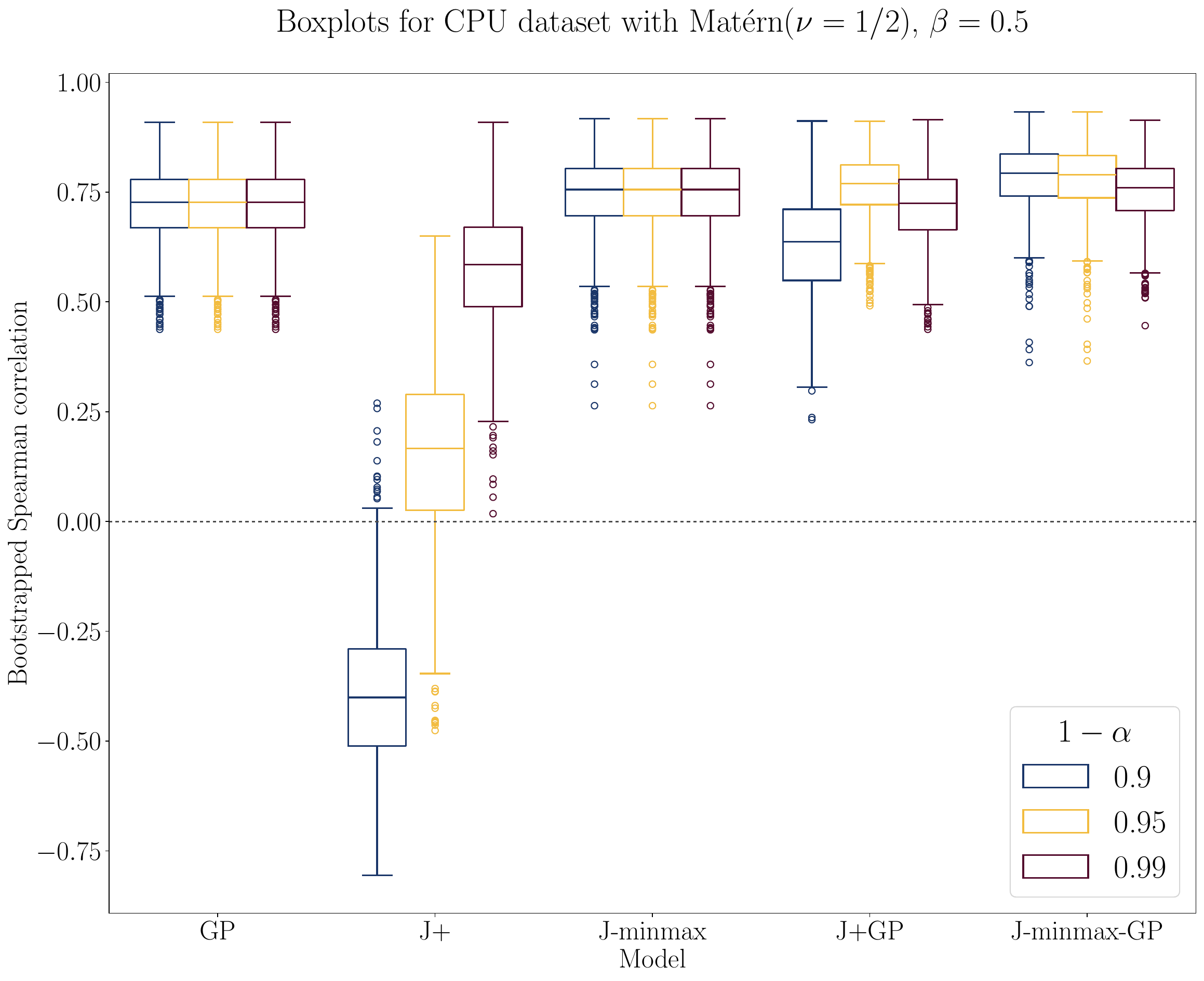}
    \caption{Boxplots of the bootstrapped Spearman correlations obtained for the different methods used to regress the CPU dataset.}
    \label{fig:cpu_boxplot}
\end{figure}

In the comprehensive results presented in Table~\ref{tab:cpu}, our J+GP method consistently achieves the smallest average width for the $90\%$ ($\nu=3/2, \beta = 1/2$) and $95\%$ ($\nu=5/2, \beta=3/2$) confidence levels. Notably, no empirical coverage is attained at the $99\%$ confidence level for any GP since the soft-threshold of $0.986$ outlined in Table~\ref{tab:cov_th} is not achieved by any of the estimators. The metamodel demonstrating the highest correlation with model precision across all coverage rates is characterized by the Matérn-$1/2$ kernel with the non-conformity score divided by a standard deviation elevated to the power of $\beta=0.5$. Figure~\ref{fig:cpu_boxplot} presents the bootstrapped Spearman correlation box-plots for the Matérn-$1/2$, $\beta=0.5$ on the different estimators. It can be seen that the J-minmax-GP method gives the highest correlation, and is closely followed by the GP credibility intervals and the standard J-minmax model. We observe that the standard Jackknife+ method provides non-adaptive intervals since the correlation obtained is significantly low. However, with the use of the posterior standard-deviation as in the J+GP estimator, we see that this method can be made more adaptive. It is interesting to point out that the standard minmax method in this setting is adaptive as well given the high Spearman correlation value observed. 

\begin{table}[htbp]
\setlength{\tabcolsep}{3pt}
  \centering
  \caption{CPU dataset. Empirical coverage rate, average width and Spearman correlation for different predictive intervals (standard Bayesian credibility, cross-conformal and the proposed estimator) for different Matérn kernels and for three confidence levels. In red and underlined: lowest widths and highest Spearman correlations obtained under the soft-coverage condition described in Table~\ref{tab:cov_th}.}\label{tab:cpu}%
    \footnotesize{% Table generated by Excel2LaTeX from sheet 'CPU'
\begin{tabular}{ccc|rrr|rrr|rrr}
\multirow{2}[1]{*}{Method} & \multicolumn{1}{c}{\multirow{2}[1]{*}{Matérn}} & \multicolumn{1}{c|}{\multirow{2}[1]{*}{$\beta$}} & \multicolumn{3}{c|}{Coverage} & \multicolumn{3}{c|}{Average width} & \multicolumn{3}{c}{Spearman corr.} \\
      &       &       & 90\%  & 95\%  & 99\%  & 90\%  & 95\%  & 99\%  & 90\%  & 95\%  & 99\% \\
\midrule
\multicolumn{1}{c}{\multirow{3}[2]{*}{GP credibility intervals}} & 1/2   & \multirow{3}[2]{*}{} & 0.976 & 0.976 & 0.976 & 91.640 & 109.195 & 143.507 & 0.720 & 0.720 & 0.720 \\
      & 3/2   &       & 0.976 & 0.976 & 0.976 & 72.942 & 86.916 & 114.226 & 0.492 & 0.492 & 0.492 \\
      & 5/2   &       & 0.976 & 0.976 & 0.976 & 70.899 & 84.481 & 111.027 & 0.626 & 0.626 & 0.626 \\
\midrule
\multicolumn{1}{c}{\multirow{3}[2]{*}{J+}} & 1/2   & \multirow{3}[2]{*}{} & 0.905 & 0.905 & 0.976 & 29.753 & 40.752 & 358.625 & -0.390 & 0.155 & 0.574 \\
      & 3/2   &       & 0.952 & 0.952 & 0.976 & 30.446 & 51.964 & 278.951 & 0.089 & 0.136 & \multicolumn{1}{l}{n.c} \\
      & 5/2   &       & 0.952 & 0.952 & 0.976 & 32.203 & 51.954 & 267.514 & -0.250 & -0.403 & \multicolumn{1}{l}{n.c} \\
\midrule
\multicolumn{1}{c}{\multirow{3}[2]{*}{J-minmax}} & 1/2   & \multirow{3}[2]{*}{} & 0.929 & 0.929 & 0.976 & 48.858 & 59.762 & 374.038 & 0.743 & 0.743 & 0.743 \\
      & 3/2   &       & 0.952 & 0.952 & 0.976 & 47.645 & 69.265 & 296.378 & 0.434 & 0.434 & 0.434 \\
      & 5/2   &       & 0.952 & 0.952 & 0.976 & 48.934 & 69.355 & 284.805 & 0.511 & 0.511 & 0.511 \\
\midrule
\multirow{9}[6]{*}{J+GP} & \multirow{3}[2]{*}{1/2} & 0.5   & 0.905 & 0.905 & 0.976 & 28.142 & 38.662 & 268.953 & 0.627 & 0.762 & 0.717 \\
      &       & 1     & 0.881 & 0.905 & 0.976 & \textcolor[rgb]{ 1,  0,  0}{\underline{25.390}} & 36.992 & 204.518 & 0.741 & 0.749 & 0.717 \\
      &       & 1.5   & 0.881 & 0.905 & 0.976 & 25.983 & 32.639 & 157.892 & 0.684 & 0.731 & 0.717 \\
\cmidrule{2-12}      & \multirow{3}[2]{*}{3/2} & 0.5   & 0.952 & 0.952 & 0.976 & 29.936 & 48.053 & 204.434 & 0.485 & 0.383 & 0.502 \\
      &       & 1     & 0.952 & 0.952 & 0.976 & 30.056 & 42.484 & 151.826 & 0.487 & 0.464 & 0.500 \\
      &       & 1.5   & 0.952 & 0.952 & 0.976 & 29.616 & 39.036 & 114.987 & 0.508 & 0.467 & 0.500 \\
\cmidrule{2-12}      & \multirow{3}[2]{*}{5/2} & 0.5   & 0.952 & 0.952 & 0.976 & 32.125 & 48.999 & 198.370 & 0.533 & 0.376 & 0.632 \\
      &       & 1     & 0.952 & 0.952 & 0.976 & 31.355 & 42.255 & 148.834 & 0.607 & 0.606 & 0.635 \\
      &       & 1.5   & 0.952 & 0.952 & 0.976 & 29.587 & \textcolor[rgb]{ 1,  0,  0}{\underline{38.569}} & 113.974 & 0.573 & 0.592 & 0.635 \\
\midrule
\multirow{9}[6]{*}{J-minmax-GP} & \multirow{3}[2]{*}{1/2} & 0.5   & 0.929 & 0.952 & 0.976 & 47.310 & 57.670 & 285.721 & \textcolor[rgb]{ 1,  0,  0}{\underline{0.782}} & \textcolor[rgb]{ 1,  0,  0}{\underline{0.779}} & 0.751 \\
      &       & 1     & 0.929 & 0.952 & 0.976 & 44.336 & 56.077 & 221.599 & 0.774 & 0.776 & 0.747 \\
      &       & 1.5   & 0.952 & 0.952 & 0.976 & 45.221 & 51.342 & 174.991 & 0.775 & 0.758 & 0.745 \\
\cmidrule{2-12}      & \multirow{3}[2]{*}{3/2} & 0.5   & 0.952 & 0.952 & 0.976 & 47.156 & 65.644 & 221.724 & 0.470 & 0.484 & 0.504 \\
      &       & 1     & 0.952 & 0.952 & 0.976 & 47.376 & 59.539 & 168.792 & 0.488 & 0.473 & 0.501 \\
      &       & 1.5   & 0.952 & 0.952 & 0.976 & 47.207 & 56.278 & 131.438 & 0.488 & 0.479 & 0.498 \\
\cmidrule{2-12}      & \multirow{3}[2]{*}{5/2} & 0.5   & 0.952 & 0.952 & 0.976 & 49.031 & 66.470 & 215.538 & 0.543 & 0.564 & 0.614 \\
      &       & 1     & 0.952 & 0.952 & 0.976 & 48.284 & 58.817 & 165.922 & 0.578 & 0.594 & 0.627 \\
      &       & 1.5   & 0.952 & 0.952 & 0.976 & 45.840 & 54.843 & 130.651 & 0.596 & 0.607 & 0.624 \\
\bottomrule
\end{tabular}%
}
\end{table}%

%%-----------------------------%%
\subsection{Benchmark on UQ analytical functions}
\label{sec:morokoff}
%%-----------------------------%%

The second example is the Morokoff \& Caflisch function \citep{morokoff_caflisch_1995} defined on the unit cube $[0, 1]^{d}$ by: 
\begin{equation}
    g(X) = \frac{1}{2}\left(1 + \frac{1}{d}\right)^{d}\prod_{i=1}^{d} (X^{(i)})^{1/d}.
\end{equation}
We choose $d=10$ and use $N=600$ samples drawn according to the multivariate normal distribution $\mathcal{N}(0,\bm{C})$ with the variance-covariance matrix $\bm{C}$ described in \citep{acharki_robust_2023}.

 The output of the function is perturbed with an additive noise following a zero-mean normal distribution with standard deviation $\sigma^{2} = 10^{-4}$. Here we use a $75\%-25\%$ split of the database for training and testing respectively and an empirical nugget $\sigma_{\epsilon}=10^{-4}$. 
 
 As seen in Table~\ref{tab:q2_mse}, for all three regularity parameters the predictivity coefficient is high with only a minor variation between $\nu=3/2$ and $\nu=5/2$. Here again, our method is of interest for completing the GP quality assessment. It outperforms the standard estimators on most soft coverage rates threshold outlined in Table~\ref{tab:cov_th}. As can be seen in Table~\ref{tab:morokoff}, the J+GP method with Matérn-$3/2$ and $\beta=1$ gives the smallest average width for the $95\%$ coverage rate and the Matérn-$5/2$ with a power $\beta=1$ produces the smallest widths for $1-\alpha = 99\%$. The Spearman correlation between the widths and the approximation error is the most significant with the J-minmax-GP estimator on all coverage rates with the Matérn-$5/2$ and $\beta=0.5$ and $1$ respectively~(see Figure~\ref{fig:morokoff_caflisch_corr}). Moreover, a comparison is described the in Appendix~\ref{secA3}, where another GP model is fitted with a nugget $\sigma_{\epsilon} = 0.1$. We show that, while this metamodel achieves almost the same $Q^2$ and MSE as our well-specified model, its correlations are significantly smaller. This could be explained by the fact that the approximation error of the misspecified model is driven essentially by the noise (as it is not well captured by the GP). This example highlights the importance of examining the correlation between the errors of the metamodel and the width of the prediction intervals. In the Appendix~\ref{secA3}, we treat the case of an additional analytical function, namely the wing-weight function \citep{forrester_2008} and our methodology achieves similar results.

\begin{table}[htbp]
  \setlength{\tabcolsep}{1.75pt}
  \centering
  \caption{Noisy Morokoff \& Caflisch analytical function. Empirical coverage rate, average width and Spearman correlation for different predictive intervals (standard Bayesian credibility, cross-conformal and the proposed estimator) for different Matérn kernels and for three confidence levels. In red and underlined: lowest widths and highest Spearman correlations obtained under the soft-coverage condition described in Table~\ref{tab:cov_th}.}\label{tab:morokoff}%  
% Table generated by Excel2LaTeX from sheet 'Morokoff_10_4'
\begin{tabular}{ccc|rrr|rrr|rrr}
\multirow{2}[1]{*}{Method} & \multicolumn{1}{c}{\multirow{2}[1]{*}{Matérn}} & \multicolumn{1}{c|}{\multirow{2}[1]{*}{$\beta$}} & \multicolumn{3}{c|}{Coverage} & \multicolumn{3}{c|}{Average width} & \multicolumn{3}{c}{Spearman corr.} \\
      &       &       & 90\%  & 95\%  & 99\%  & 90\%  & 95\%  & 99\%  & 90\%  & 95\%  & 99\% \\
\midrule
\multicolumn{1}{c}{\multirow{3}[2]{*}{GP credibility intervals}} & 1/2   & \multirow{3}[2]{*}{} & 0.933 & 0.967 & 0.983 & 0.166 & 0.198 & 0.260 & 0.119 & 0.119 & 0.119 \\
      & 3/2   &       & 0.908 & 0.933 & 0.958 & \textcolor[rgb]{ 1,  0,  0}{\underline{0.127}} & 0.151 & 0.198 & 0.213 & 0.213 & 0.213 \\
      & 5/2   &       & 0.858 & 0.917 & 0.958 & 0.117 & 0.139 & 0.183 & 0.253 & 0.253 & 0.253 \\
\midrule
\multicolumn{1}{c}{\multirow{3}[2]{*}{J+}} & 1/2   & \multirow{3}[2]{*}{} & 0.875 & 0.942 & 0.992 & 0.138 & 0.177 & 0.319 & -0.061 & -0.118 & 0.075 \\
      & 3/2   &       & 0.917 & 0.950 & 0.983 & 0.143 & 0.183 & 0.312 & -0.067 & -0.260 & -0.085 \\
      & 5/2   &       & 0.908 & 0.958 & 0.983 & 0.142 & 0.180 & 0.292 & -0.203 & -0.167 & 0.091 \\
\midrule
\multicolumn{1}{c}{\multirow{3}[2]{*}{J-minmax}} & 1/2   & \multirow{3}[2]{*}{} & 0.908 & 0.958 & 0.992 & 0.154 & 0.194 & 0.337 & 0.068 & 0.068 & 0.068 \\
      & 3/2   &       & 0.958 & 0.975 & 0.992 & 0.168 & 0.210 & 0.340 & 0.208 & 0.208 & 0.208 \\
      & 5/2   &       & 0.958 & 0.958 & 0.992 & 0.173 & 0.208 & 0.321 & 0.266 & 0.266 & 0.266 \\
\midrule
\multirow{9}[6]{*}{J+GP} & \multirow{3}[2]{*}{1/2} & 0.5   & 0.883 & 0.942 & 0.992 & 0.138 & 0.173 & 0.316 & 0.117 & 0.125 & 0.162 \\
      &       & 1     & 0.883 & 0.942 & 0.983 & 0.138 & 0.172 & 0.302 & 0.111 & 0.125 & 0.131 \\
      &       & 1.5   & 0.867 & 0.933 & 0.983 & 0.134 & 0.169 & 0.284 & 0.115 & 0.122 & 0.118 \\
\cmidrule{2-12}      & \multirow{3}[2]{*}{3/2} & 0.5   & 0.900 & 0.967 & 0.983 & 0.133 & 0.173 & 0.274 & 0.208 & 0.225 & 0.206 \\
      &       & 1     & 0.917 & 0.958 & 0.983 & 0.131 & \textcolor[rgb]{ 1,  0,  0}{\underline{0.167}} & 0.259 & 0.213 & 0.220 & 0.211 \\
      &       & 1.5   & 0.900 & 0.950 & 0.975 & 0.133 & 0.172 & 0.256 & 0.215 & 0.211 & 0.206 \\
\cmidrule{2-12}      & \multirow{3}[2]{*}{5/2} & 0.5   & 0.883 & 0.967 & 0.983 & 0.134 & 0.177 & 0.261 & 0.245 & 0.247 & 0.271 \\
      &       & 1     & 0.917 & 0.958 & 0.975 & 0.139 & 0.174 & 0.262 & 0.261 & 0.259 & 0.257 \\
      &       & 1.5   & 0.917 & 0.942 & 0.967 & 0.145 & 0.183 & 0.249 & 0.256 & 0.248 & 0.257 \\
\midrule
\multirow{9}[6]{*}{J-minmax-GP} & \multirow{3}[2]{*}{1/2} & 0.5   & 0.900 & 0.958 & 0.992 & 0.154 & 0.189 & 0.334 & 0.126 & 0.132 & 0.146 \\
      &       & 1     & 0.917 & 0.950 & 0.992 & 0.154 & 0.188 & 0.319 & 0.145 & 0.148 & 0.150 \\
      &       & 1.5   & 0.908 & 0.950 & 0.983 & 0.150 & 0.184 & 0.300 & 0.148 & 0.150 & 0.147 \\
\cmidrule{2-12}      & \multirow{3}[2]{*}{3/2} & 0.5   & 0.950 & 0.975 & 0.983 & 0.158 & 0.196 & 0.300 & 0.259 & 0.264 & 0.264 \\
      &       & 1     & 0.950 & 0.967 & 0.983 & 0.157 & 0.192 & 0.283 & 0.259 & 0.260 & 0.253 \\
      &       & 1.5   & 0.942 & 0.975 & 0.983 & 0.158 & 0.197 & 0.280 & 0.256 & 0.251 & 0.241 \\
\cmidrule{2-12}      & \multirow{3}[2]{*}{5/2} & 0.5   & 0.950 & 0.967 & 0.983 & 0.162 & 0.208 & 0.287 & \textcolor[rgb]{ 1,  0,  0}{\underline{0.300}} & \textcolor[rgb]{ 1,  0,  0}{\underline{0.306}} & 0.301 \\
      &       & 1     & 0.950 & 0.958 & 0.992 & 0.166 & 0.203 & \textcolor[rgb]{ 1,  0,  0}{\underline{0.290}} & 0.298 & 0.292 & \textcolor[rgb]{ 1,  0,  0}{\underline{0.285}} \\
      &       & 1.5   & 0.950 & 0.967 & 0.975 & 0.174 & 0.213 & 0.274 & 0.288 & 0.282 & 0.277 \\
\bottomrule
\end{tabular}%

\end{table}%

\begin{figure}[!ht]
    \centering
\includegraphics[width=0.9\textwidth]{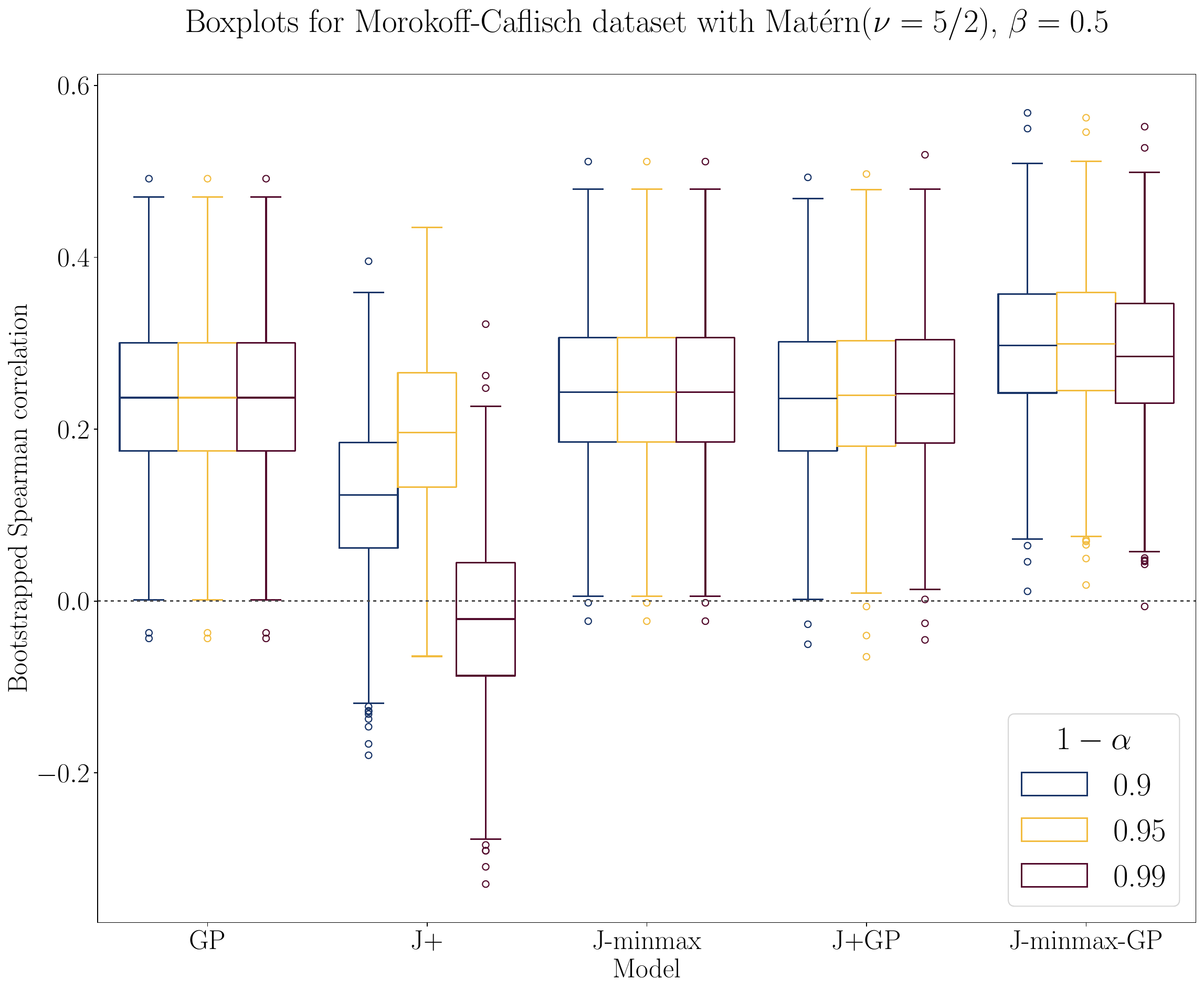}
    \caption{Boxplots of the bootstrapped Spearman correlations obtained for the different methods used to regress the noisy Morokoff \& Caflisch function.}
    \label{fig:morokoff_caflisch_corr}
\end{figure}

%%-----------------------------%%
\subsection{Benchmark on an industrial use case: the THYC-Puffer-DEPOTHYC code}
%%-----------------------------%%

The following industrial test case is linked to the issue of clogging in steam generators (SG) of pressurized water nuclear reactors. Over time, SGs of some reactors may face the challenge of clogging, a deposition phenomenon that heightens the risk of mechanical and vibration stress on tube bundles and internal structures. Additionally, it affects their response during hypothetical accidental transients. For robustifying maintenance planning, EDF R\&D has developed a numerical multi-physics computational chain named ``THYC-Puffer-DEPOTHYC'' (TPD). The numerical tool utilizes specific physical models to replicate the kinetics of clogging, generating time-dependent clogging rate profiles for specific SGs. Some input parameters of this code are subject to uncertainties. For better understanding the sensitivity of the output uncertainty with respect to the input variability, a full methodology using polynomial chaos expansion surrogate models and advanced global sensitivity techniques has been proposed in \citep{jaber_et_al_2023}. Here, we dispose of a database of $1000$ Monte Carlo simulations, with $7$ features to predict the clogging rate at a specific time. The input distribution of the features is recalled in Table~\ref{tab:tpd_variables}. More information about the physical nature of the features can be found in \citep{jaber_et_al_2023}.

\begin{figure}[!ht]
    \centering
\includegraphics[width=0.9\textwidth]{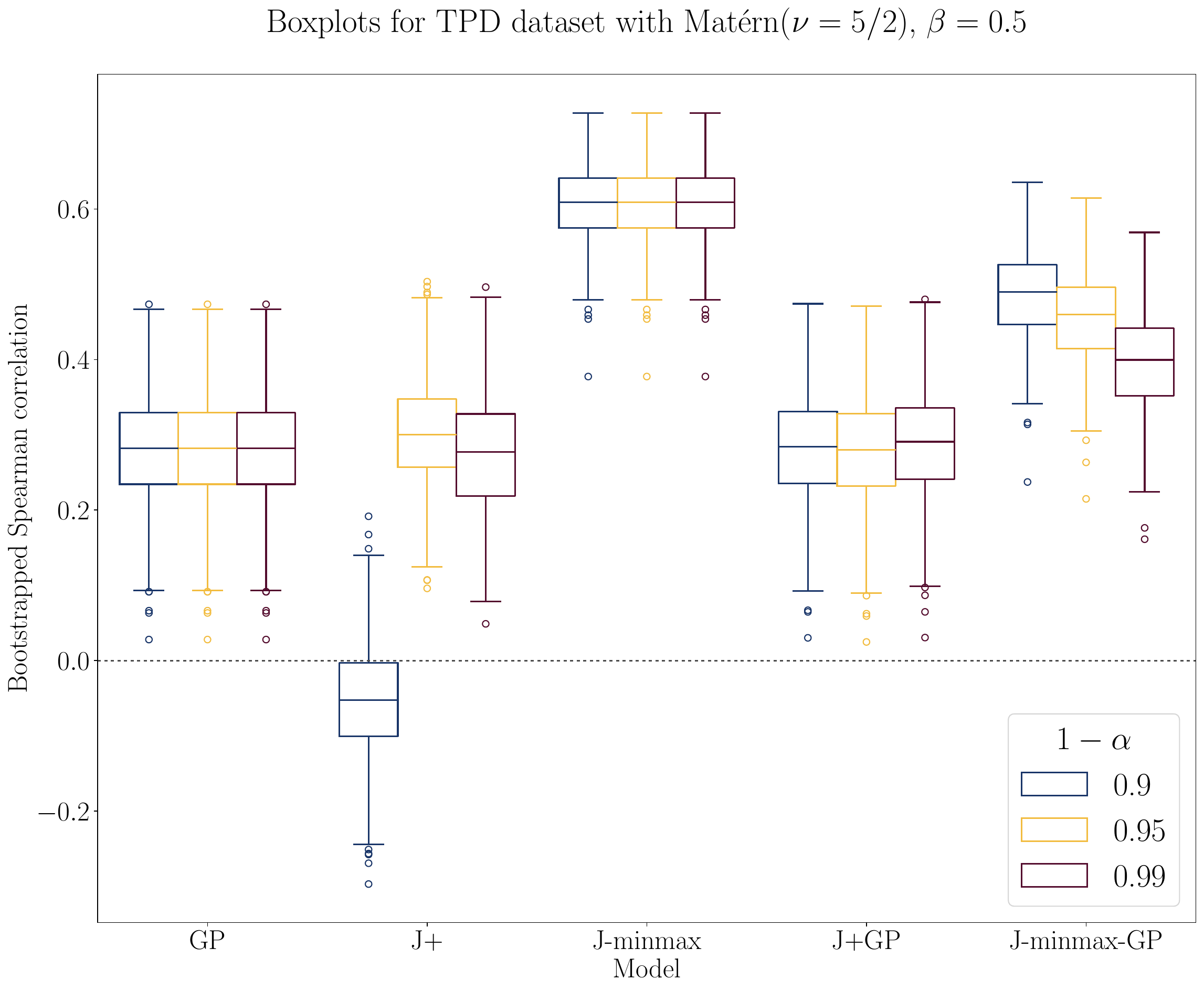}
    \caption{Boxplots of the Spearman correlations obtained for the different methods used to regress the THYC-Puffer-DEPOTHYC industrial testcase.}
    \label{fig:cpu_corr}
\end{figure}

The outcomes of the analysis are detailed in Table~\ref{tab:dthyc}. The predictive capability of the posterior GP proves exceedingly high ($Q^{2} \geq 0.99$) for all regularity parameters, making it challenging to pinpoint the optimal candidate. In the pursuit of establishing which leads to a robust GP metamodel of THYC-Puffer-DEPOTHYC to speed-up industrial studies on clogging, the different conformal predictors reveal an advantage for a GP employing Matérn-$3/2$ and Matérn-$5/2$ prior kernels. Specifically, across all soft-coverage rates, the J-minmax-GP intervals for $\nu=5/2$ exhibit the smallest average widths, while Bayesian credibility intervals fall short of meeting the required threshold at this regularity parameter. Notably, the standard J-minmax estimator demonstrates a noteworthy level of adaptivity, particularly for the coverage rates of $90\%$ and $95\%$, surpassing correlations obtained with the Bayesian credibility interval widths. This is most visible in Figure~\ref{fig:cpu_corr} where the minmax method for $\nu=5/2$ achieves the highest median Spearman correlation among all the other methods. It should be emphasized however that the minmax method is more conservative and our J-minmax-GP estimator produces smaller intervals for the same coverage rates.
Additionally, the J-minmax-GP estimator displays the highest correlation at the soft-$99\%$ level for the different Matérn priors that is $\nu=3/2$.
Therefore in view of applications for speeding-up industrial uncertainty studies of clogging, one can view the GP metamodel with zero mean and Matérn-$3/2$ or $5/2$ kernel optimized through MLE as the best candidate for the metamodeling of THYC-Puffer-DEPOTHYC. 

\begin{table}[htbp]
    \centering
    \caption{Distributions of the input variables of THYC-Puffer-DEPOTHYC.}
    \label{tab:tpd_variables}
    \begin{tabular}{cccc}
        \hline
        \textbf{Component} & \textbf{Distribution} & \textbf{Component} & \textbf{Distribution} \\
        \hline
        $X^{(1)}$ & $\mathcal{N}(101.6, 4.0)$ & $X^{(5)}$ & $\mathcal{T}(0.5, 5.0, 10.0) \times 10^{-6}$ \\
        $X^{(2)}$ & $\mathcal{N}(0.0233, 0.0005)$ & $X^{(6)}$ & $\mathcal{T}(1.0, 4.5, 8.0) \times 10^{-9}$ \\
        $X^{(3)}$ & $\mathcal{T}(0.2, 0.3, 0.5)$ & $X^{(7)}$ & $\mathcal{T}(0.1, 7.8, 12)\times 10^{-4}$ \\
        $X^{(4)}$ & $\mathcal{T}(0.01, 0.05, 0.3)$ & & \\
        \hline
\end{tabular} 
\end{table}

\begin{table}[htbp]
  \centering
  \setlength{\tabcolsep}{3pt}
  \caption{THYC-Puffer-DEPOTHYC industrial dataset. Empirical coverage rate, average width and Spearman correlation for different predictive intervals (standard Bayesian credibility, cross-conformal and the proposed estimator) for different Matérn kernels and for three confidence levels. In red and underlined: lowest widths and highest Spearman correlations obtained under the soft-coverage condition described in Table~\ref{tab:cov_th}.}
    % Table generated by Excel2LaTeX from sheet 'DTHYC'
\begin{tabular}{ccc|rrr|rrr|rrr}
\multirow{2}[1]{*}{Method} & \multicolumn{1}{c}{\multirow{2}[1]{*}{Matérn}} & \multicolumn{1}{c|}{\multirow{2}[1]{*}{$\beta$}} & \multicolumn{3}{c|}{Coverage} & \multicolumn{3}{c|}{Average width} & \multicolumn{3}{c}{Spearman corr.} \\
      &       &       & 90\%  & 95\%  & 99\%  & 90\%  & 95\%  & 99\%  & 90\%  & 95\%  & 99\% \\
\midrule
\multicolumn{1}{c}{\multirow{3}[2]{*}{GP credibility intervals}} & 1/2   & \multirow{3}[2]{*}{} & 0.960 & 0.975 & 0.975 & 4.717 & 5.621 & 7.387 & 0.463 & 0.463 & 0.463 \\
      & 3/2   &       & 0.915 & 0.940 & 0.950 & 2.000 & 2.384 & 3.133 & 0.353 & 0.353 & 0.353 \\
      & 5/2   &       & 0.850 & 0.885 & 0.945 & 1.632 & 1.944 & 2.555 & 0.281 & 0.281 & 0.281 \\
\midrule
\multicolumn{1}{c}{\multirow{3}[2]{*}{J+}} & 1/2   & \multirow{3}[2]{*}{} & 0.855 & 0.900 & 0.975 & 2.438 & 3.610 & 7.391 & 0.266 & -0.223 & 0.132 \\
      & 3/2   &       & 0.840 & 0.905 & 0.975 & 1.529 & 2.031 & 3.943 & -0.355 & 0.043 & 0.202 \\
      & 5/2   &       & 0.840 & 0.920 & 0.965 & 1.353 & 1.836 & 3.109 & -0.052 & 0.301 & 0.273 \\
\midrule
\multicolumn{1}{c}{\multirow{3}[2]{*}{J-minmax}} & 1/2   & \multirow{3}[2]{*}{} & 0.860 & 0.920 & 0.975 & 2.763 & 3.943 & 7.711 & 0.666 & 0.666 & 0.666 \\
      & 3/2   &       & 0.890 & 0.920 & 0.980 & 1.857 & 2.350 & 4.260 & \textcolor[rgb]{ 1,  0,  0}{\underline{0.653}} & 0.653 & 0.653 \\
      & 5/2   &       & 0.905 & 0.950 & 0.980 & 1.763 & 2.233 & 3.505 & 0.606 & \textcolor[rgb]{ 1,  0,  0}{\underline{0.606}} & 0.606 \\
\midrule
\multirow{9}[6]{*}{J+GP} & \multirow{3}[2]{*}{1/2} & 0.5   & 0.840 & 0.885 & 0.975 & 2.346 & 3.396 & 6.889 & 0.463 & 0.450 & 0.456 \\
      &       & 1     & 0.845 & 0.895 & 0.975 & 2.314 & 3.198 & 6.367 & 0.469 & 0.466 & 0.458 \\
      &       & 1.5   & 0.835 & 0.895 & 0.975 & 2.284 & 3.047 & 6.126 & 0.465 & 0.462 & 0.464 \\
\cmidrule{2-12}      & \multirow{3}[2]{*}{3/2} & 0.5   & 0.835 & 0.910 & 0.975 & 1.438 & 2.002 & 3.565 & 0.364 & 0.338 & 0.348 \\
      &       & 1     & 0.840 & 0.925 & 0.955 & 1.523 & 2.058 & 3.215 & 0.351 & 0.345 & 0.352 \\
      &       & 1.5   & 0.865 & 0.925 & 0.970 & 1.702 & 2.270 & 3.773 & 0.356 & 0.353 & 0.355 \\
\cmidrule{2-12}      & \multirow{3}[2]{*}{5/2} & 0.5   & 0.855 & 0.910 & 0.960 & 1.326 & 1.765 & 3.113 & 0.283 & 0.279 & 0.289 \\
      &       & 1     & 0.845 & 0.905 & 0.970 & 1.509 & 2.072 & 3.689 & 0.279 & 0.280 & 0.288 \\
      &       & 1.5   & 0.855 & 0.925 & 0.980 & 2.046 & 2.959 & 5.909 & 0.280 & 0.279 & 0.282 \\
\midrule
\multirow{9}[6]{*}{J-minmax-GP} & \multirow{3}[2]{*}{1/2} & 0.5   & 0.870 & 0.925 & 0.975 & 2.674 & 3.732 & 7.228 & 0.660 & 0.638 & 0.584 \\
      &       & 1     & 0.870 & 0.920 & 0.975 & 2.638 & 3.523 & 6.700 & 0.617 & 0.592 & 0.546 \\
      &       & 1.5   & 0.865 & 0.910 & 0.975 & 2.612 & 3.376 & 6.460 & 0.588 & 0.568 & 0.529 \\
\cmidrule{2-12}      & \multirow{3}[2]{*}{3/2} & 0.5   & 0.885 & 0.945 & 0.985 & 1.753 & 2.335 & 3.897 & 0.596 & 0.558 & 0.489 \\
      &       & 1     & 0.900 & 0.950 & 0.985 & 1.852 & 2.387 & 3.543 & 0.519 & 0.489 & 0.449 \\
      &       & 1.5   & 0.920 & 0.950 & 0.995 & 2.022 & 2.592 & 4.094 & 0.482 & 0.456 & \textcolor[rgb]{ 1,  0,  0}{\underline{0.421}} \\
\cmidrule{2-12}      & \multirow{3}[2]{*}{5/2} & 0.5   & 0.910 & 0.955 & 0.985 & \textcolor[rgb]{ 1,  0,  0}{\underline{1.730}} & \textcolor[rgb]{ 1,  0,  0}{\underline{2.178}} & 3.503 & 0.487 & 0.456 & 0.397 \\
      &       & 1     & 0.895 & 0.960 & 0.995 & 1.918 & 2.477 & \textcolor[rgb]{ 1,  0,  0}{\underline{4.080}} & 0.424 & 0.390 & 0.349 \\
      &       & 1.5   & 0.915 & 0.970 & 0.995 & 2.451 & 3.370 & 6.312 & 0.383 & 0.357 & 0.322 \\
\bottomrule
\end{tabular}%

  \label{tab:dthyc}%
\end{table}%

%%-----------------------------%%
\subsection{Synthesis of the results}
%%-----------------------------%%

We have shown that for a given confidence level, the study of the average width of the prediction intervals and their Spearman correlation with the error enhances the evaluation of the metamodel quality.
This has been numerically exemplified with standard ML datasets, analytical UQ functions and a complex computer-code industrial case where the selection of different metamodels, for instance through different Matérn kernels on the sole basis of the $Q^{2}$ is not fully reliable. Indeed, our cross-conformal J+GP and J-minmax-GP as well as the standard J-minmax conformal estimators usually give the smallest prediction intervals width under the soft-empirical coverage condition and achieve better correlations than the standard Bayesian credibility intervals, however good the $Q^2$ is.

%%==================================%%
\section{Conclusion and perspectives}
\label{sec:conclusion}
%%==================================%%

In this work, we explore the idea of conformalizing GPs in the cross-conformal prediction paradigm in order to robustify the GP metamodel evaluation for industrial applications. The idea is to make more reliable GPs prediction in a context in which we are not sure that our surrogate model is well specified. To this end, we adapt the classical LOO non-conformity score by weighting it with the local posterior standard deviation of the GP raised to different powers. This method allows the conformal predictive intervals to endow a better adaptivity, thus having a varying interval span for different new test points. Moreover, our J+GP prediction interval enjoys the same theoretical marginal coverage property as the Jackknife+ and its min-max variant. In order to quantify this adaptivity of the confidence interval, we evaluate the Spearman correlation between the width of the intervals and the absolute value of the metamodel local error. We exhibit that our methods achieve a better adaptivity than both standard cross-conformal methods and GPs credibility intervals. We demonstrate the ability of our methodology for GP model selection between  different prior regularity parameters for the Matérn kernels. Moreover, we show how the proposed methodology can help in assessing the validity of a GP metamodel for industrial applications through a real use case related to nuclear engineering. In the case of very noisy data, additional work is necessary for optimizing the nugget parameter of the GP prior in order to achieve better correlations between the model precision and the constructed predictive intervals. Future line of research would be to generalize this methodology to families of deterministic metamodels like polynomial chaos expansions which do not come equipped with an inherent stochastic structure such as the GPs or to more general statistical models that come with a quantifiable notion of dispersion.

%%%%%%%%%%%%%%%%%%%%%%%%%%%%%%%%%%%%%%%%%%%%%%%%%%%%%%%%%%%%%%%%%%
\begin{appendices}

%%==================================%%
\section{Proof of theorem~\ref{th:theorem_1}}
\label{secA1}
%%==================================%%

\begin{proof}
Assume that:
\begin{equation}
    Y = g(X) + \epsilon,
\end{equation}
with $\epsilon$ representing noise and that a statistical learning model $\widehat{g}$ is trained on the database $\mathcal{D}_{n} = \{(X_{i}, Y_{i})\}_{i=1}^{n}$. Let $(X_{n+1}, Y_{n+1}) \in \mathcal{X} \times \mathcal{Y}$ be a new point. We denote by $\mathcal{D}_{n+1}:= \mathcal{D} \cup \{ (X_{n+1},Y_{n+1}) \}$.
Let $\widehat{g}_{-(i, j)}\; \forall i \neq j \in \{1, ..., n+1\}$, be the statistical learning $\mathcal{D}_{n+1}\backslash \{(X_{i},Y_{i}), (X_{j},Y_{j})\}$. By exchangeability we have that $\widehat{g}_{-(i,j)} = \widehat{g}_{-(j,i)}$ and $\widehat{g}_{-i} = \widehat{g}_{-(i,n+1)}$. Let us denote by $\widehat{\sigma}(X_{i})$ an estimator of the standard-deviation of $\widehat{g}$ and assume $\widehat{\sigma} > 0$ and similarly for the corresponding LOO. Similarly as for the Gaussian non-conformity score we define:
\begin{equation}
    R^{\textnormal{LOO}\sigma}_{i} = \frac{\lvert Y_{i} - \widehat{g}(X_{i})\rvert}{\widehat{\sigma}_{-i}^{\;\beta}(X_{i})},
\end{equation}
we then proceed and define $R\in \mathcal{M}_{n+1}(\mathbb{R})$ as:
\begin{equation}
R_{ij} = \begin{cases}
         + \infty & \text{if } i = j, \\
       \lvert Y_{i} - \widehat{g}_{-(i, j)}(X_{i}) \rvert \;/\; \widehat{\sigma}_{-(i,j)}^{\;\beta}(X_{i}) & \text{if } i \neq j,
        \end{cases}
\end{equation}
For simplifying the notations we will now fix $\beta=1$. We proceed in defining the matrix $A\in\mathcal{M}_{n+1}(\{0,1\})$:
\begin{equation}
    A_{ij} = \mathds{1}\{R_{ij} > R_{ji}\}. 
\end{equation}
It can be easily observed that $A_{ij} = 1 \Leftrightarrow A_{ji} = 0$. The strange set associated to $A$ for $\alpha \in (0,1)$ is:
\begin{equation}
    \mathcal{S}(A):= \left\{i\in \{1,\ldots,n+1\}\;:\; \sum_{j=1}^{n+1}A_{ij} \geq (1-\alpha)(n+1) \right\},
\end{equation}
in other words, a point $i$ is \emph{strange} if the residual $R_{ij}$ compared with $R_{ji}$ for all $j \neq i$ is larger for a given fraction of comparisons. 

We start by bounding the cardinal of $\mathcal{S}(A)$. Let $i$ be a strange point. This point can lose against \emph{at most} $\alpha(n+1) - 1$ other strange points, since it has to win at least $(1-\alpha)(n+1)$ times and it cannot win against himself. Let $s = \lvert\mathcal{S}(A) \rvert$, we now group pairs of strange points by the losing point. There are at most $s$ possibilities for the loser and for each one, it can lose at most $\alpha(n+1) - 1$ times. Thus there are at most $s\times(\alpha(n+1) - 1)$ \emph{pairs} of strange points.

We can now bound the number of ways we can choose two points in $\mathcal{S}(A)$ and we obtain:
\begin{equation}
    \frac{s(s-1)}{2}\leq s\times(\alpha(n+1) - 1),
\end{equation}
and rearranging:
\begin{equation}
    s \leq 2\alpha(n+1).
\end{equation}
We assume that the dataset $\mathcal{D}_{n+1}$ is exchangeable. Thus, using permutation matrices $\Pi$ that maps a $j\in\{1,\ldots,n+1\}$ to $n+1$ (such that $\Pi_{j, n+1} = 1$), we prove that: 
\begin{equation}
    \mathbb{P}(n+1\in\mathcal{S}(A)) = \mathbb{P}(j\in\mathcal{S}(\Pi A \Pi^{T})) = \mathbb{P}(j\in\mathcal{S}(A)).
\end{equation}
Therefore, any point is equally likely to be strange. We have then: 
\begin{equation}
    \mathbb{P}(n+1 \in \mathcal{S}(A)) = \frac{1}{n+1}\sum_{j = 1}^{n+1} \mathbb{P}(j \in \mathcal{S}(A)) = \frac{\mathbb{E}[ \lvert\mathcal{S}(A)\rvert ]}{n+1} \leq 2\alpha.
\end{equation}
We can now reconnect with the definition of prediction intervals. Let us suppose that $Y_{n+1}\notin \widehat{C}^{*}_{n,\alpha}
$. Then, for at least $(1-\alpha)(n+1)$ values $i$ in $\{1,\ldots,n+1\}$, we have: 
\begin{equation}
    Y_{n+1} > \widehat{g}_{-i}(X_{n+1}) + R_{i}^{\textnormal{LOO}\sigma}\times \widehat{\sigma}_{-i}(X_{n+1}),
\end{equation}
or,
\begin{equation}
    Y_{n+1} < \widehat{g}_{-i}(X_{n+1}) - R_{i}^{\textnormal{LOO}\sigma}\times \widehat{\sigma}_{-i}(X_{n+1}).
\end{equation}
Finally we can compute:
\begin{equation*}
\begin{aligned}
(1-\alpha)(n+1) &\leq \sum_{i=1}^{n+1} \mathds{1}\left\{Y_{n+1} \notin \widehat{g}_{-i}(X_{n+1}) \pm R_{i}^{\textnormal{LOO}\sigma}\times \widehat{\sigma}_{-i}(X_{n+1})\right\} \\
& = \sum_{i=1}^{n+1} \mathds{1}\left\{
R_{i}^{\textnormal{LOO}\sigma} \times \widehat{\sigma}_{-i}(X_{n+1})< | Y_{n+1} - \widehat{g}_{-i}(X_{n+1}) |
\right\}
\\
& = \sum_{i=1}^{n+1} \mathds{1}\left\{R_{i}^{\textnormal{LOO}\sigma} < \frac{\lvert Y_{n+1} - \widehat{g}_{-i}(X_{n+1})\rvert}{\widehat{\sigma}_{-i}(X_{n+1})}\right\}\\
&=\sum_{i=1}^{n+1}\mathds{1}\left\{ \frac{\lvert Y_{i} - \widehat{g}_{-i}(X_{i}) \rvert}{\widehat{\sigma}_{-i}(X_{i})} < \frac{\lvert Y_{n+1} - \widehat{g}_{-i}(X_{n+1})\rvert}{\widehat{\sigma}_{-i}(X_{n+1})}\right\}\\
&= \sum_{i=1}^{n+1} \mathds{1}\left\{ R_{i,n+1} < R_{n+1,i}\right\} = \sum_{i=1}^{n+1} A_{n+1,i}
\end{aligned}
\end{equation*}
where the last equality above is obtained with the identities  $\widehat{\sigma}_{-i}(X_{i}) = \widehat{\sigma}_{-(i,n+1)}(X_{i})$ and $\widehat{g}_{-i}(X_{i}) = \widehat{g}_{-(i,n+1)}(X_{i})$. Therefore $n+1\in \mathcal{S}(A)$ and: 
\begin{equation}
    \mathbb{P} \left(g(X_{n+1}) \notin \widehat{C}_{n,\alpha}^{*}(X_{n+1}) \right) \leq \mathbb{P}(n+1\in\mathcal{S}(A)) \leq 2\alpha.
\end{equation}
\end{proof}

%%==================================%%
\section{Proof of theorem~\ref{th:theorem_2}}
\begin{proof}
    Assume the same hypothesis as in the previous theorem and we make use of the same definitions and notations. We define the matrix $\widetilde{R}\in \mathcal{M}_{n+1}(\mathbb{R})$ as:
\begin{equation}
\widetilde{R}_{ij} = \begin{cases}
         + \infty & \text{if } i = j, \\
       R_{ij} \times \widehat{\sigma}_{-(i,j)}(X_{n+1}) & \text{if } i \neq j,
        \end{cases}
\end{equation}

    We redefine the matrix $A\in\mathcal{M}_{n+1}(\{0,1\})$:
    \begin{equation}
        A_{ij} = \mathds{1}\{\min_{j^{'}} \widetilde{R}_{ij^{'}} \geq \widetilde{R}_{ji}\},
    \end{equation}
    where  $\min_{j^{'}} \widetilde{R}_{ij^{'}}$ is the smallest residual for the point $i$ when leaving out any point $j^{'}\in\{1,\ldots,n\}$. We start by bound the number of strange points, choose:
    \begin{equation}
        i_{*} \in \argmin_{i\in\mathcal{S}(A)}\min_{j^{'}}\widetilde{R}_{ij^{'}}.
    \end{equation}
    We can observe that for all strange point $j\in\mathcal{S}(A)$, the matrix element $A_{i_{*}j}$ is null. Indeed this comes since by definition:
    \begin{equation}
        \forall j\in\mathcal{S}(A), \; \widetilde{R}_{ji_{*}} \geq \min_{j^{'}}\widetilde{R}_{jj^{'}} \geq \min_{j^{'}} \widetilde{R}_{i_{*}{j^{'}}}.
    \end{equation}
    We can then easily bound the number of strange points using that $i_{*}\in\mathcal{S}(A)$:
    \begin{equation}
        n+1 - |\mathcal{S}(A)| \geq \sum_{j=1}^{n+1} A_{i_{*}j} \geq (1-\alpha)(n+1),
    \end{equation}
    and a rearrangement gives:
    \begin{equation}
        |\mathcal{S}(A)| \leq \alpha(n+1).
    \end{equation}
    Using the exchangeability property in the same fashion as the preceding proof we have that:
    \begin{equation}
        \mathbb{P}\left(n+1 \in \mathcal{S}(A)\right) \leq \alpha.
    \end{equation}
    
     Let us suppose now that $Y_{n+1}\notin \widehat{C}^{*-minmax}_{n,\alpha}
$. Then, for at least $(1-\alpha)(n+1)$ values $i$ in $\{1,\ldots,n+1\}$, we have: 
\begin{equation}
    Y_{n+1} > \max_{i=1,\ldots,n}\widehat{g}_{-i}(X_{n+1}) + R_{i}^{\textnormal{LOO}\sigma}\times \widehat{\sigma}_{-i}(X_{n+1}),
\end{equation}
or,
\begin{equation}
    Y_{n+1} < \min_{i=1,\ldots,n}\widehat{g}_{-i}(X_{n+1}) - R_{i}^{\textnormal{LOO}\sigma}\times \widehat{\sigma}_{-i}(X_{n+1}).
\end{equation}
We denote $\widehat{g}_{\min}(X_{i}):= \min_{j=1\ldots,n} \widehat{g}_{-j}(X_{i})$ and similarly for $\widehat{g}_{\max}$ and $\widetilde{R}_{i}(X_{n+1}):= R_{i}^{\textnormal{LOO}\sigma}\times\widehat{\sigma}_{-i}(X_{n+1})$. Finally we can compute:
\begin{equation*}
\begin{aligned}
(1-\alpha)(n+1) &\leq \sum_{i=1}^{n+1} \mathds{1}\left\{Y_{n+1} \notin \left[ \widehat{g}_{\min}(X_{n+1}) - \widetilde{R}_{i}(X_{n+1}),  \widehat{g}_{\max}(X_{n+1}) + \widetilde{R}_{i}(X_{n+1})\right]\right\} \\
&= \sum_{i=1}^{n+1}\mathds{1}\left\{\min_{j=1,\ldots,n} \lvert Y_{n+1} - \widehat{g}_{-j}(X_{n+1})\rvert \geq R_{i}^{\textnormal{LOO}\sigma}\times \widehat{\sigma}_{-i}(X_{n+1})\right\}\\
&= \sum_{i=1}^{n+1}\mathds{1}\left\{\min_{j=1,\ldots,n} \frac{\lvert Y_{n+1} - \widehat{g}_{-j}(X_{n+1})\rvert}{\widehat{\sigma}_{-j}(X_{n+1})}\times \widehat{\sigma}_{-j}(X_{n+1}) \geq R_{i}^{\textnormal{LOO}\sigma}\times \widehat{\sigma}_{-i}(X_{n+1})\right\}\\
&= \sum_{i=1}^{n+1}\mathds{1}\left\{\min_{j=1,\ldots,n}\frac{\lvert Y_{n+1} - \widehat{g}_{-(n+1,j)}(X_{n+1})\rvert}{\widehat{\sigma}_{-(n+1,j)}(X_{n+1})} \times \widehat{\sigma}_{-(n+1,j)}(X_{n+1}) \geq \right. \\
&\hspace{2em}\qquad \left.  \frac{\lvert Y_{i} - \widehat{g}_{-(i,n+1)}(X_{i}) \rvert}{\widehat{\sigma}_{-(i,n+1)}(X_{i})} \times\widehat{\sigma}_{-(i,n+1)}(X_{n+1})\right\}\\
&= \sum_{i=1}^{n+1}\mathds{1}\left\{\min_{j=1,\ldots,n} R_{n+1,j}\times \widehat{\sigma}_{-(n+1,j)}(X_{n+1}) \geq R_{i,n+1}\times \widehat{\sigma}_{-(i,n+1)}(X_{n+1})\right\}\\
&= \sum_{i=1}^{n+1}\mathds{1}\left\{\min_{j=1,\ldots,n} \widetilde{R}_{n+1,j}\geq \widetilde{R}_{i,n+1}\right\}\\
&= \sum_{i=1}^{n+1} A_{n+1,i}
\end{aligned}
\end{equation*}

Therefore $n+1 \in \mathcal{S}(A)$ and we conclude as in the preceding theorem.
\end{proof}
\label{secA2}
%%==================================%%

%%==================================%%
\section{Additional numerical results}
\label{secA3}
%%==================================%%
Predictive performance of the GPs as well as a description of the dataset used is available in Table~\ref{tab:q2_mse_annex}

\begin{table}[htbp]
\centering
  \caption{Summary of the performance metrics of the GP metamodels of additional datasets.}
  \setlength{\tabcolsep}{7pt}
  \begin{tabular}{c|c||c|c|c|c}
Matérn &  & MPG & Wing-Weight & Morokoff\&Caflisch with $\sigma_{\epsilon}=10^{-1}$\\
\hline
\shortstack{$\;$ \\ $\;$ } & \shortstack{$d$ \\ $N$ \\ $\%\text{train}$ \\ $\%\text{test}$ \\
$\sigma_{\epsilon}$} & \shortstack{$7$ \\ $392$\\ $80$ \\ $20$ \\ $0.1$} & \shortstack{$7$ \\ $600$\\ $75$ \\ $25$ \\ $0.1$} & \shortstack{$10$ \\ $600$\\ $75$ \\ $25$ \\ $10^{-1}$}  \\
\hline
\shortstack{$1/2$ \\ $\;$ } & \shortstack{$Q^{2}$ \\ MSE} & \shortstack{$0.893$\\ $5.45$} & \shortstack{$0.980$ \\ $55.15$} & \shortstack{$0.918$ \\ $2.5\times10^{-3}$} \\
\hline
\shortstack{$3/2$ \\ $\;$} & \shortstack{$Q^{2}$ \\ MSE} & \shortstack{$0.892$\\ $5.49$} & \shortstack{$0.984$ \\ $43.33$} & \shortstack{$0.93$ \\ $2.11\times10^{-3}$}\\
\hline
\shortstack{$5/2$ \\ $\;$} & \shortstack{$Q^{2}$ \\ MSE} & \shortstack{$0.890$\\ $5.58$} & \shortstack{$0.984$ \\ $41.69$} & \shortstack{$0.932$ \\ $2.06\times10^{-3}$}\\
\hline
\end{tabular}

  \label{tab:q2_mse_annex}%
\end{table}

\begin{table}[htbp]
\centering
  \caption{Threshold above which it is established that the coverage rate is achieved for different confidence levels.}
  \setlength{\tabcolsep}{7pt}
   \begin{tabular}{c|c|c|c|}
Dataset & 90\% & 95\% & 99\% \\
\hline
MPG & 0.880 & 0.937 & 0.983 \\
Wing-Weight & 0.882 & 0.938 & 0.985 \\
\hline
\end{tabular}

  \label{tab:cov_th_annex}%
\end{table}

%%%%%%%%%%%
\subsection{MPG Dataset}
%%%%%%%%%%%
The dataset on miles per gallon \citep{misc_auto_mpg_9} is a regression database containing $398$ entries. Its goal is to predict fuel consumption using $7$ features, which include both continuous and discrete attributes. After removing the undefined lines, the database is reduced to a size of $392$ entries. Each kernel employs $80\%$ of the original dataset for training and $20\%$ for testing and computing various indices and intervals. For the nugget-effect we take a standard deviation of $0.1$.

\begin{figure}[!ht]
    \centering
\includegraphics[width=1.0\textwidth]{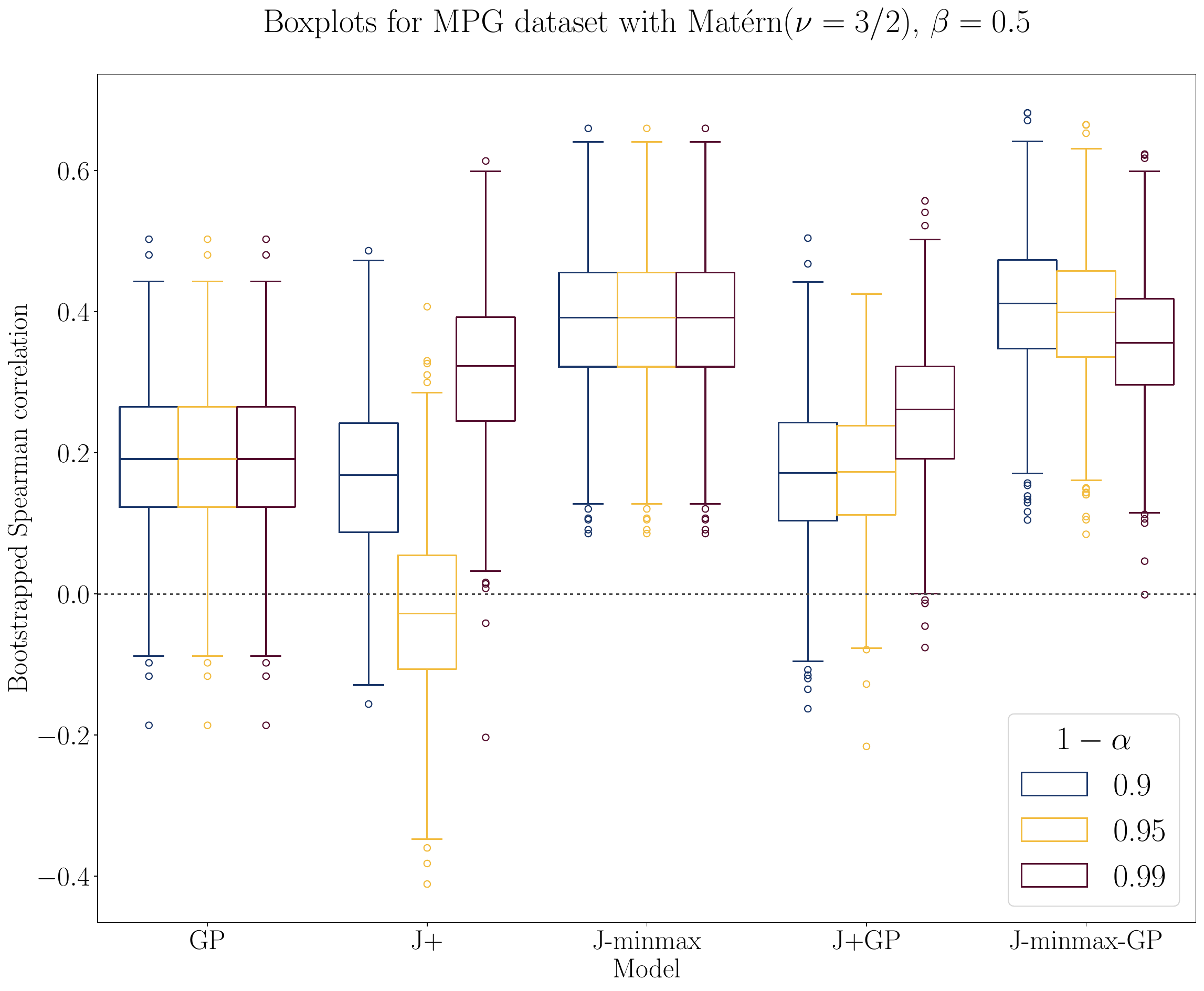}
    \caption{Boxplots of the bootstrapped Spearman correlations obtained for the different methods used to regress the MPG dataset.}
    \label{fig:mpg}
\end{figure}

The results are displayed in Table~\ref{tab:mpg} and the correlation boxplots in Figure~\ref{fig:mpg}. Across the three kernels, the GPs demonstrate strong performance in terms of $Q^{2}$. In this initial test, our J+GP method stands out by achieving the smallest average width for a Matérn-$5/2$ kernel at two confidence levels ($95\%$ and $99\%$) with a $\beta$-power set to 1. Notably, the minmax method exhibits the highest correlation with the model error when $\beta = 1/2$ for the 90\% and 95\% level confidence. It is worth mentioning the remarkably low correlation observed in standard GP credibility intervals and classical CP in this scenario.

\begin{table}[htbp]
  \centering
  \setlength{\tabcolsep}{3pt}
  \caption{MPG dataset. Empirical coverage rate, average width and Spearman correlation for different predictive intervals (standard Bayesian credibility, cross-conformal and the proposed estimator) for different Matérn kernels and for three confidence levels. In red and underlined: lowest widths and highest Spearman correlations obtained under the soft-coverage condition described in Table~\ref{tab:cov_th}.}
  % Table generated by Excel2LaTeX from sheet 'MPG'
\begin{tabular}{ccc|rrr|rrr|rrr}
\multirow{2}[1]{*}{Method} & \multicolumn{1}{c}{\multirow{2}[1]{*}{Matérn}} & \multicolumn{1}{c|}{\multirow{2}[1]{*}{$\beta$}} & \multicolumn{3}{c|}{Coverage} & \multicolumn{3}{c|}{Average width} & \multicolumn{3}{c}{Spearman corr.} \\
      &       &       & 90\%  & 95\%  & 99\%  & 90\%  & 95\%  & 99\%  & 90\%  & 95\%  & 99\% \\
\midrule
\multicolumn{1}{c}{\multirow{3}[2]{*}{GP credibility intervals}} & 1/2   & \multirow{3}[2]{*}{} & 0.924 & 0.949 & 1.000 & 8.882 & 10.584 & 13.910 & 0.076 & 0.074 & 0.079 \\
      & 3/2   &       & 0.937 & 0.949 & 1.000 & 8.654 & 10.312 & \textcolor[rgb]{ 1,  0,  0}{\underline{13.552}} & 0.188 & 0.188 & 0.194 \\
      & 5/2   &       & 0.911 & 0.949 & 1.000 & 8.661 & 10.320 & 13.563 & 0.211 & 0.207 & 0.211 \\
\midrule
\multicolumn{1}{c}{\multirow{3}[2]{*}{J+}} & 1/2   & \multirow{3}[2]{*}{} & 0.911 & 0.949 & 1.000 & 8.202 & 10.773 & 19.337 & -0.199 & 0.021 & 0.131 \\
      & 3/2   &       & 0.911 & 0.937 & 1.000 & 8.017 & 10.756 & 19.764 & 0.166 & -0.026 & 0.309 \\
      & 5/2   &       & 0.911 & 0.949 & 1.000 & 8.081 & 10.777 & 19.716 & 0.139 & -0.193 & 0.130 \\
\midrule
\multicolumn{1}{c}{\multirow{3}[2]{*}{J-minmax}} & 1/2   & \multirow{3}[2]{*}{} & 0.949 & 0.975 & 1.000 & 9.280 & 11.842 & 20.386 & 0.401 & 0.400 & \textcolor[rgb]{ 1,  0,  0}{\underline{0.398}} \\
      & 3/2   &       & 0.949 & 0.962 & 1.000 & 8.951 & 11.830 & 20.701 & 0.384 & 0.383 & 0.380 \\
      & 5/2   &       & 0.949 & 0.962 & 1.000 & 8.986 & 11.765 & 20.577 & 0.398 & 0.388 & 0.394 \\
\midrule
\multirow{9}[6]{*}{J+GP} & \multirow{3}[2]{*}{1/2} & 0.5   & 0.899 & 0.949 & 1.000 & 8.147 & 10.810 & 18.590 & 0.101 & 0.074 & 0.130 \\
      &       & 1     & 0.886 & 0.949 & 1.000 & 8.179 & 11.193 & 19.519 & 0.081 & 0.061 & 0.075 \\
      &       & 1.5   & 0.886 & 0.949 & 1.000 & 8.242 & 11.092 & 20.564 & 0.077 & 0.083 & 0.053 \\
\cmidrule{2-12}      & \multirow{3}[2]{*}{3/2} & 0.5   & 0.911 & 0.937 & 1.000 & \textcolor[rgb]{ 1,  0,  0}{\underline{7.946}} & 10.801 & 19.019 & 0.166 & 0.173 & 0.260 \\
      &       & 1     & 0.899 & 0.949 & 1.000 & 7.976 & 10.783 & 19.354 & 0.196 & 0.239 & 0.229 \\
      &       & 1.5   & 0.899 & 0.949 & 1.000 & 8.113 & 10.594 & 19.771 & 0.188 & 0.235 & 0.207 \\
\cmidrule{2-12}      & \multirow{3}[2]{*}{5/2} & 0.5   & 0.911 & 0.949 & 1.000 & 8.022 & 10.669 & 18.935 & 0.220 & 0.082 & 0.330 \\
      &       & 1     & 0.911 & 0.949 & 1.000 & 8.041 & 10.546 & 19.030 & 0.216 & 0.202 & 0.262 \\
      &       & 1.5   & 0.899 & 0.949 & 1.000 & 8.071 & 10.541 & 19.239 & 0.218 & 0.245 & 0.249 \\
\midrule
\multirow{9}[6]{*}{J-minmax-GP} & \multirow{3}[2]{*}{1/2} & 0.5   & 0.949 & 0.975 & 1.000 & 9.176 & 11.873 & 19.588 & 0.364 & 0.334 & 0.270 \\
      &       & 1     & 0.962 & 0.962 & 1.000 & 9.221 & 12.337 & 20.596 & 0.296 & 0.244 & 0.187 \\
      &       & 1.5   & 0.962 & 0.962 & 1.000 & 9.275 & 12.138 & 21.703 & 0.241 & 0.192 & 0.156 \\
\cmidrule{2-12}      & \multirow{3}[2]{*}{3/2} & 0.5   & 0.949 & 0.962 & 1.000 & 8.957 & 11.833 & 19.889 & \textcolor[rgb]{ 1,  0,  0}{\underline{0.404}} & 0.398 & 0.345 \\
      &       & 1     & 0.949 & 0.962 & 1.000 & 8.963 & 11.748 & 20.302 & 0.374 & 0.369 & 0.297 \\
      &       & 1.5   & 0.949 & 0.962 & 1.000 & 9.120 & 11.490 & 20.753 & 0.350 & 0.337 & 0.276 \\
\cmidrule{2-12}      & \multirow{3}[2]{*}{5/2} & 0.5   & 0.949 & 0.962 & 1.000 & 8.942 & 11.671 & 19.633 & 0.399 & \textcolor[rgb]{ 1,  0,  0}{\underline{0.402}} & 0.375 \\
      &       & 1     & 0.949 & 0.962 & 1.000 & 8.988 & 11.539 & 19.853 & 0.385 & 0.386 & 0.333 \\
      &       & 1.5   & 0.949 & 0.962 & 1.000 & 8.970 & \textcolor[rgb]{ 1,  0,  0}{\underline{11.429}} & 20.100 & 0.376 & 0.358 & 0.311 \\
\bottomrule
\end{tabular}%

  \label{tab:mpg}%
\end{table}%

%%%%%%%%%%%
\subsection{Wing-weight function}
%%%%%%%%%%%
The wing-weight function has been proposed in \citep{forrester_2008}. It is an analytic function with input dimension $10$ and scalar output that is used for UQ benchmarks. If we denote $X = (X^{(1)}, \ldots, X^{(10)})$, the function is given by:
\begin{equation}
\begin{aligned}
    g(X) = & 0.036(X^{(1)})^{0.758}(X^{(2)})^{0.0035}\left(\frac{X^{(3)}}{\cos^{2}(X^{(4)})}\right)^{0.6}(X^{(5)})^{0.006} \\
    & \times (X^{(6)})^{0.04}\left(\frac{100X^{(7)}}{\cos(X^{(4)})}\right)^{-0.3}(X^{(8)}X^{(9)})^{0.49} + X^{(1)}X^{(10)}.
\end{aligned}
\end{equation}
The response variable $Y$ is generated according to the following formula: 
\begin{equation}
    Y_i = g(X_i) + \epsilon_i,
\end{equation}
where $X_{i} = (X^{(1)}_{i}, \ldots, X^{(10)}_{i})$ for all $i$ has its domain in Table~\ref{tab:ww_variables} and the $\epsilon_i$ are i.i.d. samples from a normal distribution with zero mean and $\sigma^{2} = 25$ as is done in \citep{acharki_robust_2023}. We build a sample of $n=600$ observations. The GP is estimated using a nugget effect with an observation noise variance set to $0.1$. 

\begin{figure}[!ht]
    \centering
\includegraphics[width=1.0\textwidth]{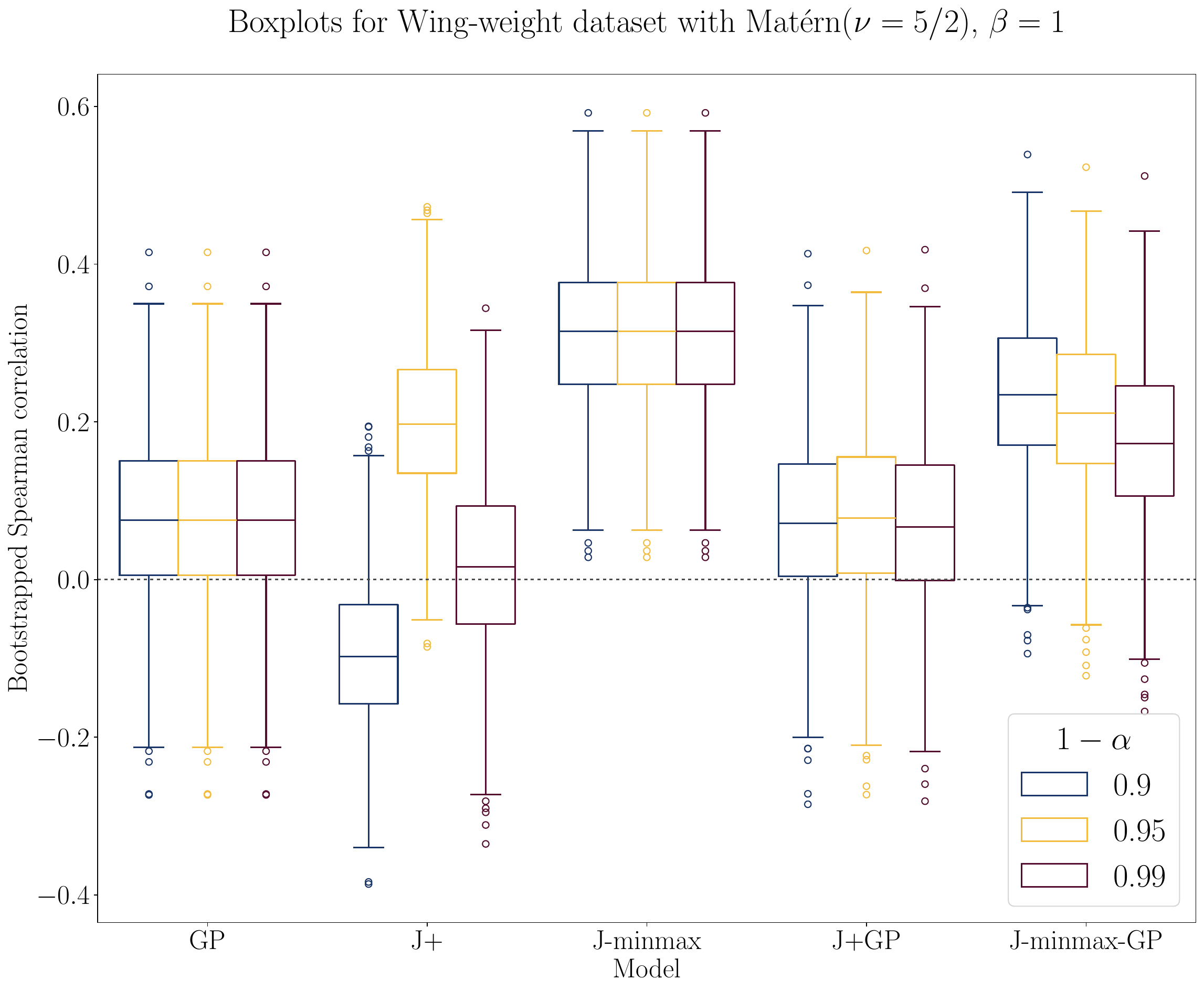}
    \caption{Boxplots of the bootstrapped Spearman correlations obtained for the different methods used to regress the noisy Wing-weight function.}
    \label{fig:ww}
\end{figure}

In the results presented in Table~\ref{tab:wing_weigth_noisy} and Figure~\ref{fig:ww}, it shows that the credibility intervals of the GP are larger than those of the conformal methods hence giving an empirical coverage higher than the desired one. For the $90\%$ and $99\%$ coverage, our method J-minmax-GP has a lower prediction intervals width than the other methods. Here the highest correlations are achieved for the J-minmax methods, however this correlation is very low. One possible reason is that we have a function with a high noise and that most of the errors of the model are purely aleatory.

\begin{table}[htbp]
    \centering
    \caption{Domains of the input variables of the wing-weight function.}
    \begin{tabular}{cccc}
        \hline
        \textbf{Component} & \textbf{Domain} & \textbf{Component} & \textbf{Domain} \\
        \hline
        $X^{(1)}$ & $[150, 200]$ & $X^{(6)}$ & $[0.5, 1]$ \\
        $X^{(2)}$ & $[220, 300]$ & $X^{(7)}$ & $[0.08, 0.18]$ \\
        $X^{(3)}$ & $[6, 10]$ & $X^{(8)}$ & $[2.5, 6]$ \\
        $X^{(4)}$ & $[-10, 10]$ & $X^{(9)}$ & $[1700, 2500]$ \\
        $X^{(5)}$ & $[16, 45]$ & $X^{(10)}$ & $[0.025, 0.08]$ \\
        \hline
\end{tabular}
    \label{tab:ww_variables}
\end{table}

\begin{table}[htbp]
  \centering
\setlength{\tabcolsep}{1.75pt}
  \caption{Wing-weight function with noise. Empirical coverage rate, average width and Spearman correlation for different predictive intervals (standard Bayesian credibility, cross-conformal and the proposed estimator) for different Matérn kernels and for three confidence levels. In red and underlined: lowest widths and highest Spearman correlations obtained under the soft-coverage condition described in Table~\ref{tab:cov_th_annex}.}
% Table generated by Excel2LaTeX from sheet 'Wing-weight noisy'
\begin{tabular}{ccc|rrr|rrr|rrr}
\multirow{2}[1]{*}{Method} & \multicolumn{1}{c}{\multirow{2}[1]{*}{Matérn}} & \multicolumn{1}{c|}{\multirow{2}[1]{*}{$\beta$}} & \multicolumn{3}{c|}{Coverage} & \multicolumn{3}{c|}{Average width} & \multicolumn{3}{c}{Spearman corr.} \\
      &       &       & 90\%  & 95\%  & 99\%  & 90\%  & 95\%  & 99\%  & 90\%  & 95\%  & 99\% \\
\midrule
\multicolumn{1}{c}{\multirow{3}[2]{*}{GP credibility intervals}} & 1/2   & \multirow{3}[2]{*}{} & 0.983 & 0.992 & 1.000 & 35.317 & 42.082 & 55.306 & -0.007 & -0.007 & -0.007 \\
      & 3/2   &       & 0.958 & 0.975 & 1.000 & 25.816 & 30.762 & 40.428 & -0.051 & -0.051 & -0.051 \\
      & 5/2   &       & 0.933 & 0.975 & 1.000 & 24.202 & 28.838 & 37.900 & -0.064 & -0.064 & -0.064 \\
\midrule
\multicolumn{1}{c}{\multirow{3}[2]{*}{J+}} & 1/2   & \multirow{3}[2]{*}{} & 0.883 & 0.958 & 0.992 & 23.281 & 28.764 & 39.213 & -0.117 & 0.035 & 0.074 \\
      & 3/2   &       & 0.883 & 0.942 & 0.983 & \textcolor[rgb]{ 1,  0,  0}{\underline{21.070}} & 24.792 & 32.899 & \textcolor[rgb]{ 1,  0,  0}{\underline{0.118}} & -0.143 & 0.093 \\
      & 5/2   &       & 0.867 & 0.958 & 0.983 & 20.500 & 24.706 & 32.205 & -0.054 & -0.095 & 0.012 \\
\midrule
\multicolumn{1}{c}{\multirow{3}[2]{*}{J-minmax}} & 1/2   & \multirow{3}[2]{*}{} & 0.917 & 0.967 & 0.992 & 25.336 & 30.983 & 41.291 & 0.074 & \textcolor[rgb]{ 1,  0,  0}{\underline{0.074}} & \textcolor[rgb]{ 1,  0,  0}{\underline{0.074}} \\
      & 3/2   &       & 0.942 & 0.958 & 0.983 & 23.249 & 27.092 & 35.029 & 0.018 & 0.018 & 0.018 \\
      & 5/2   &       & 0.917 & 0.958 & 0.983 & 22.479 & 26.761 & 34.029 & 0.016 & 0.016 & 0.016 \\
\midrule
\multirow{9}[6]{*}{J+GP} & \multirow{3}[2]{*}{1/2} & 0.5   & 0.908 & 0.942 & 0.992 & 23.487 & 28.151 & 39.879 & -0.012 & 0.000 & -0.025 \\
      &       & 1     & 0.892 & 0.942 & 0.992 & 23.449 & 27.560 & 40.092 & -0.022 & 0.020 & 0.003 \\
      &       & 1.5   & 0.892 & 0.950 & 0.992 & 23.350 & 27.079 & 41.230 & -0.014 & -0.005 & -0.005 \\
\cmidrule{2-12}      & \multirow{3}[2]{*}{3/2} & 0.5   & 0.875 & 0.942 & 0.983 & 20.375 & 24.500 & 34.076 & -0.045 & -0.051 & -0.047 \\
      &       & 1     & 0.867 & 0.942 & 0.992 & 20.051 & 24.850 & 35.298 & -0.044 & -0.055 & -0.049 \\
      &       & 1.5   & 0.842 & 0.942 & 0.983 & 19.882 & 25.463 & 34.698 & -0.045 & -0.046 & -0.051 \\
\cmidrule{2-12}      & \multirow{3}[2]{*}{5/2} & 0.5   & 0.875 & 0.942 & 0.983 & 20.399 & \textcolor[rgb]{ 1,  0,  0}{\underline{24.188}} & 33.088 & -0.076 & -0.069 & -0.060 \\
      &       & 1     & 0.875 & 0.933 & 0.983 & 20.141 & 24.173 & 32.515 & -0.083 & -0.076 & -0.062 \\
      &       & 1.5   & 0.867 & 0.942 & 0.983 & 20.195 & 24.731 & 31.841 & -0.071 & -0.066 & -0.060 \\
\midrule
\multirow{9}[6]{*}{J-minmax-GP} & \multirow{3}[2]{*}{1/2} & 0.5   & 0.933 & 0.967 & 0.992 & 25.688 & 30.306 & 41.947 & 0.058 & 0.055 & 0.051 \\
      &       & 1     & 0.933 & 0.958 & 0.992 & 25.500 & 29.608 & 42.134 & 0.052 & 0.047 & 0.042 \\
      &       & 1.5   & 0.933 & 0.950 & 0.992 & 25.449 & 29.113 & 43.360 & 0.046 & 0.043 & 0.028 \\
\cmidrule{2-12}      & \multirow{3}[2]{*}{3/2} & 0.5   & 0.900 & 0.950 & 1.000 & 22.468 & 26.647 & 36.253 & 0.000 & 0.006 & 0.001 \\
      &       & 1     & 0.900 & 0.958 & 1.000 & 22.205 & 26.914 & 37.562 & -0.007 & -0.012 & -0.022 \\
      &       & 1.5   & 0.900 & 0.950 & 1.000 & 21.969 & 27.658 & 36.865 & -0.020 & -0.021 & -0.027 \\
\cmidrule{2-12}      & \multirow{3}[2]{*}{5/2} & 0.5   & 0.900 & 0.958 & 1.000 & 22.418 & 26.007 & 35.155 & -0.006 & -0.009 & -0.018 \\
      &       & 1     & 0.900 & 0.958 & 0.992 & 22.154 & 26.101 & \textcolor[rgb]{ 1,  0,  0}{\underline{34.387}} & -0.024 & -0.029 & -0.035 \\
      &       & 1.5   & 0.900 & 0.950 & 0.983 & 22.191 & 26.725 & 33.551 & -0.035 & -0.036 & -0.040 \\
\bottomrule
\end{tabular}%

\label{tab:wing_weigth_noisy}%
\end{table}%

%%%%%%%%%%%
\subsection{Misspecified metamodel for Morokoff \& Caflisch function}
%%%%%%%%%%%

Here, we revisit the same analytical function as in Section \ref{sec:morokoff}, but instead of setting an empirical nugget $\sigma_{\epsilon} = 10^{-4}$ (equal to the actual noise of the function) we purposely use $\sigma_{\epsilon} = 10^{-1}$.

\begin{figure}[!ht]
    \centering
\includegraphics[width=1.0\textwidth]{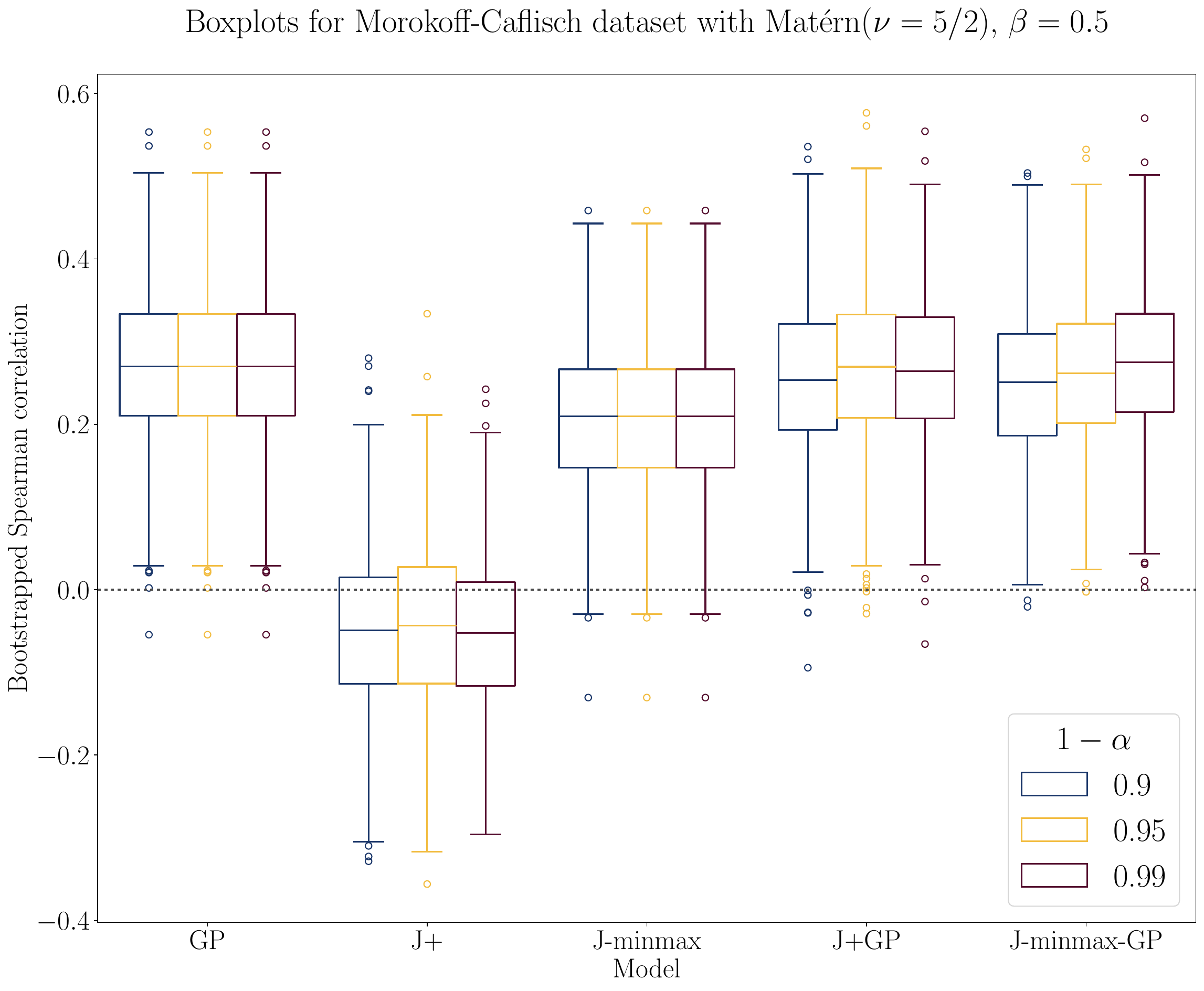}
    \caption{Boxplots of the bootstrapped Spearman correlations obtained for the different methods used to regress the noisy Morokoff \& Caflisch function with a misspecified model.}
    \label{fig:miss_morokoff}
\end{figure}

\begin{table}[!htbp]
    \setlength{\tabcolsep}{1.75pt}
    \centering
    \caption{Morokoff \& Caflisch function with $\sigma_{\epsilon}=0.1$. Empirical coverage rate, average width and Spearman correlation for different predictive intervals (standard Bayesian credibility, cross-conformal and the proposed estimator) for different Matérn kernels and for three confidence levels. In red and underlined: lowest widths and highest Spearman correlations obtained under the soft-coverage condition described in Table~\ref{tab:cov_th}.}
    \label{tab:morokoff_10_1}
% Table generated by Excel2LaTeX from sheet 'Morokoff_10_1'
\begin{tabular}{ccc|rrr|rrr|rrr}
\multirow{2}[1]{*}{Method} & \multicolumn{1}{c}{\multirow{2}[1]{*}{Matérn}} & \multicolumn{1}{c|}{\multirow{2}[1]{*}{$\beta$}} & \multicolumn{3}{c|}{Coverage} & \multicolumn{3}{c|}{Average width} & \multicolumn{3}{c}{Spearman corr.} \\
      &       &       & 90\%  & 95\%  & 99\%  & 90\%  & 95\%  & 99\%  & 90\%  & 95\%  & 99\% \\
\midrule
\multicolumn{1}{c}{\multirow{3}[2]{*}{GP credibility intervals}} & 1/2   & \multirow{3}[2]{*}{} & 0.942 & 0.967 & 0.975 & 0.181 & 0.215 & 0.283 & 0.122 & 0.122 & 0.122 \\
      & 3/2   &       & 0.933 & 0.967 & 0.983 & 0.155 & 0.185 & 0.243 & 0.115 & 0.115 & 0.115 \\
      & 5/2   &       & 0.917 & 0.967 & 0.983 & 0.151 & 0.181 & 0.237 & 0.130 & 0.130 & 0.130 \\
\midrule
\multicolumn{1}{c}{\multirow{3}[2]{*}{J+}} & 1/2   & \multirow{3}[2]{*}{} & 0.858 & 0.942 & 0.983 & 0.137 & 0.191 & 0.319 & 0.036 & 0.129 & -0.053 \\
      & 3/2   &       & 0.875 & 0.958 & 0.992 & 0.136 & 0.174 & 0.317 & -0.024 & -0.047 & 0.052 \\
      & 5/2   &       & 0.892 & 0.950 & 0.992 & 0.136 & 0.175 & 0.312 & -0.042 & -0.039 & -0.052 \\
\midrule
\multicolumn{1}{c}{\multirow{3}[2]{*}{J-minmax}} & 1/2   & \multirow{3}[2]{*}{} & 0.900 & 0.958 & 0.983 & 0.151 & 0.203 & 0.333 & 0.065 & 0.065 & 0.065 \\
      & 3/2   &       & 0.917 & 0.967 & 0.992 & 0.150 & 0.189 & 0.335 & 0.136 & 0.136 & 0.136 \\
      & 5/2   &       & 0.933 & 0.975 & 0.992 & 0.151 & 0.191 & 0.329 & 0.167 & 0.167 & 0.167 \\
\midrule
\multirow{9}[6]{*}{J+GP} & \multirow{3}[2]{*}{1/2} & 0.5   & 0.867 & 0.950 & 0.992 & 0.137 & 0.184 & 0.319 & 0.126 & 0.159 & 0.151 \\
      &       & 1     & 0.858 & 0.950 & 0.992 & 0.136 & 0.183 & 0.318 & 0.123 & 0.136 & 0.139 \\
      &       & 1.5   & 0.858 & 0.942 & 0.992 & 0.136 & 0.181 & 0.308 & 0.127 & 0.120 & 0.132 \\
\cmidrule{2-12}      & \multirow{3}[2]{*}{3/2} & 0.5   & 0.892 & 0.950 & 0.992 & 0.136 & 0.174 & \textcolor[rgb]{ 1,  0,  0}{\underline{0.305}} & 0.112 & 0.084 & 0.135 \\
      &       & 1     & 0.892 & 0.950 & 0.983 & 0.135 & 0.171 & 0.288 & 0.115 & 0.110 & 0.107 \\
      &       & 1.5   & 0.892 & 0.942 & 0.983 & 0.135 & \textcolor[rgb]{ 1,  0,  0}{\underline{0.167}} & 0.279 & 0.114 & 0.117 & 0.113 \\
\cmidrule{2-12}      & \multirow{3}[2]{*}{5/2} & 0.5   & 0.892 & 0.950 & 0.983 & 0.135 & 0.173 & 0.296 & 0.132 & 0.074 & 0.126 \\
      &       & 1     & 0.892 & 0.950 & 0.983 & 0.134 & 0.167 & 0.280 & 0.129 & 0.128 & 0.110 \\
      &       & 1.5   & 0.883 & 0.950 & 0.983 & \textcolor[rgb]{ 1,  0,  0}{\underline{0.132}} & 0.167 & 0.279 & 0.130 & 0.125 & 0.125 \\
\midrule
\multirow{9}[6]{*}{J-minmax-GP} & \multirow{3}[2]{*}{1/2} & 0.5   & 0.908 & 0.958 & 0.992 & 0.151 & 0.197 & 0.333 & 0.129 & 0.135 & 0.150 \\
      &       & 1     & 0.892 & 0.958 & 0.992 & 0.150 & 0.196 & 0.333 & 0.151 & 0.150 & 0.150 \\
      &       & 1.5   & 0.883 & 0.950 & 0.992 & 0.149 & 0.194 & 0.323 & 0.147 & 0.152 & 0.152 \\
\cmidrule{2-12}      & \multirow{3}[2]{*}{3/2} & 0.5   & 0.925 & 0.967 & 0.992 & 0.152 & 0.190 & 0.321 & 0.148 & 0.158 & 0.158 \\
      &       & 1     & 0.917 & 0.967 & 0.983 & 0.150 & 0.186 & 0.303 & 0.156 & 0.165 & 0.150 \\
      &       & 1.5   & 0.917 & 0.967 & 0.983 & 0.150 & 0.182 & 0.294 & 0.154 & 0.160 & 0.148 \\
\cmidrule{2-12}      & \multirow{3}[2]{*}{5/2} & 0.5   & 0.917 & 0.967 & 0.992 & 0.150 & 0.189 & 0.311 & 0.171 & \textcolor[rgb]{ 1,  0,  0}{\underline{0.176}} & \textcolor[rgb]{ 1,  0,  0}{\underline{0.172}} \\
      &       & 1     & 0.917 & 0.967 & 0.983 & 0.149 & 0.181 & 0.294 & \textcolor[rgb]{ 1,  0,  0}{\underline{0.174}} & 0.172 & 0.162 \\
      &       & 1.5   & 0.917 & 0.975 & 0.983 & 0.148 & 0.182 & 0.293 & 0.168 & 0.169 & 0.153 \\
\bottomrule
\end{tabular}%

\end{table}
We can see by comparing  tables \ref{tab:q2_mse_annex} and \ref{tab:q2_mse} that the $Q^2$ and MSE of metamodels with $\sigma_{\epsilon} = 10^{-1}$ and $\sigma_{\epsilon} = 10^{-4}$ respectively achieve almost the same performance; however they do not have the same correlations between the errors and the width of the prediction intervals. By comparing Table~\ref{tab:morokoff_10_1} and Figure~\ref{fig:miss_morokoff} to Table~\ref{tab:morokoff} and Figure~\ref{fig:morokoff_caflisch_corr}, we see that the correlations are almost twice higher for the well-specified model than for the misspecified one. It shows that looking at this correlation can be a good indicator.
Moreover, we can check that the highest correlation is again achieved with our proposed methodology (here J-minmax-GP). 

\end{appendices}

\newpage
\clearpage

%%==================================%%
\bibliography{sn-bibliography}%  common bib file

%%%%%%%%%%%%%%%%%%%%%%%%%%%%%%%%%%%%%%%%%%%%%%%%%%%%%%%%%%%%%%%%%%
%%%%%%%%%%%%%%%%%%%%%%%%%%%%%%%%%%%%%%%%%%%%%%%%%%%%%%%%%%%%%%%%%%
\end{document}